\documentclass[lettersize,journal]{IEEEtran}
\usepackage{amsmath,amsfonts}
\usepackage{array}
\usepackage{url}
\usepackage{graphicx}
\usepackage{cite}           
\usepackage{mathtools}
\usepackage{booktabs}
\usepackage{multirow}
\usepackage{makecell}
\usepackage{tabularx}
\usepackage{adjustbox}
\usepackage{colortbl}
\usepackage{siunitx}
\usepackage{subcaption}
\usepackage{wrapfig}
\usepackage{stfloats}       
\usepackage{xcolor}
\usepackage{enumitem}
\usepackage[normalem]{ulem}
\usepackage[hidelinks]{hyperref}
\DeclareUnicodeCharacter{0394}{\ensuremath{\Delta}}
\newcommand{\citep}[1]{\cite{#1}}

\title{EEG-PRIME: Prototype-Aligned Representation Learning with Multi-Level Conditioning for EEG Decoding}

\author{Shuailei~Zhang, \IEEEmembership{Member, IEEE},~Muyun~Jiang,~Wei~Zhang,~Jinbo~Chen, ~Zhiwei~Guo,~Yong~Li,~Yi~Ding,~and~Cuntai~Guan, \IEEEmembership{Fellow, IEEE}%
\thanks{$^{*}$S.~Zhang and M.~Jiang contributed equally to this work.}%
\thanks{$^{\dagger}$Y.~Ding and C.~Guan are the corresponding authors.}%
\thanks{S.~Zhang, M.~Jiang, ~W~Zhang,~J~Chen, Z.~Guo, Y.~Ding, and C.~Guan are with College of Computing and Data Science, Nanyang Technological University, Singapore (e-mail: ding.yi@ntu.edu.sg; ctguan@ntu.edu.sg).}%
\thanks{Cuntai Guan is also with the Centre for AI in Medicine, Nanyang Technological University, Singapore}
\thanks{Yong Li is with Southeast University, China.}}
\begin{document}

\maketitle

\begin{abstract}
Electroencephalography (EEG) decoding models often generalize poorly across datasets and subjects due to domain shifts in acquisition protocols and individual neurophysiology. 
To address these challenges, we propose EEG-PRIME, a two-stage EEG foundation model for cross-dataset multi-task decoding. EEG-PRIME combines masked pretraining with prototype-aligned instruction tuning under multi-level conditioning, enabling instruction-aware, subject-invariant EEG decoding across diverse brain-computer interface (BCI) paradigms. 
In the pretraining stage, an EEG encoder learns transferable representations via self-supervised masked reconstruction with frequency-cutoff spectral augmentation. In the instruction tuning stage, we introduce three conditioning levels: (1) a \textbf{task-semantic prompt} derived from a natural language instruction describing the decoding objective, encoded by a pre-trained sentence encoder into a fixed-dimensional embedding; (2) a \textbf{dataset-level soft embedding} that is jointly learned during training and additively combined with the task prompt to capture dataset-specific distributional characteristics; and (3) a \textbf{subject-invariance constraint} enforced via gradient reversal adversarial training, which encourages the model to suppress subject-specific variation and learn representations that generalize across individuals. The combined conditioning signal is injected into the Q-Former via \textbf{Layer-wise Query Modulation}, enabling fine-grained, layer-wise control over query representations at each transformer layer. Finally, class prototypes are defined as frozen text embeddings of category label strings, and EEG representations are matched to prototypes via cosine similarity, enabling unified prediction across heterogeneous label spaces.
Experiments on sixteen datasets spanning motor imagery, emotion recognition, attention deficit hyperactivity disorder detection, covert speech, and mental workload demonstrate consistent improvements over state-of-the-art baselines and prior EEG foundation models under cross-subject settings. Additionally, we tested EEG-PRIME on two more held-out datasets without any target-domain optimization, calibration, or linear probing, and it achieved balanced accuracy comparable to within-session calibration models, demonstrating EEG-PRIME's promising zero-shot transfer capability. Code and pretrained models are available at \url{https://github.com/ZhangShuailei/EEG-PRIME}.
\end{abstract}


\section{Introduction}
\IEEEPARstart{E}{lectroencephalography} (EEG) provides noninvasive measurements of brain dynamics with millisecond-level temporal resolution, making it particularly suitable for brain--computer interfaces (BCI) such as motor imagery (MI) decoding, emotion recognition, cognition assessment, and covert speech decoding. In practice, however, EEG decoding models often generalize poorly across datasets and subjects due to nonstationarity, low signal-to-noise ratio, and domain shifts induced by different acquisition protocols and individual neurophysiology \citep{edelman2024non}. These challenges motivate EEG foundation models (EEG-FMs) that leverage large-scale pretraining to learn transferable representations. For example, EEGPT \citep{wang2024eegpt} applies transformer-based pretraining to capture temporal dependencies, LaBraM \citep{jiang2024large} leverages masked autoencoding on large EEG corpora, and CBraMod \citep{wang2024cbramod} focuses on cross-brain modeling to facilitate cross-subject transfer.

Recent efforts further explore integrating language supervision into EEG-FMs. NeuroLM \citep{jiang2024neurolm}, for instance, aligns EEG and language embeddings by learning a text-aligned neural tokenizer and performing instruction tuning. However, existing approaches still struggle to (i)~leverage language as a controllable conditioning signal for cross-paradigm EEG decoding, (ii)~reconcile heterogeneous datasets that differ in class sets and recording characteristics without dataset-specific classifier heads, and (iii)~suppress inter-subject neurophysiological variability to learn subject-invariant representations.

\begin{figure*}[t]
    \centering
    \includegraphics[width=1.0\textwidth]{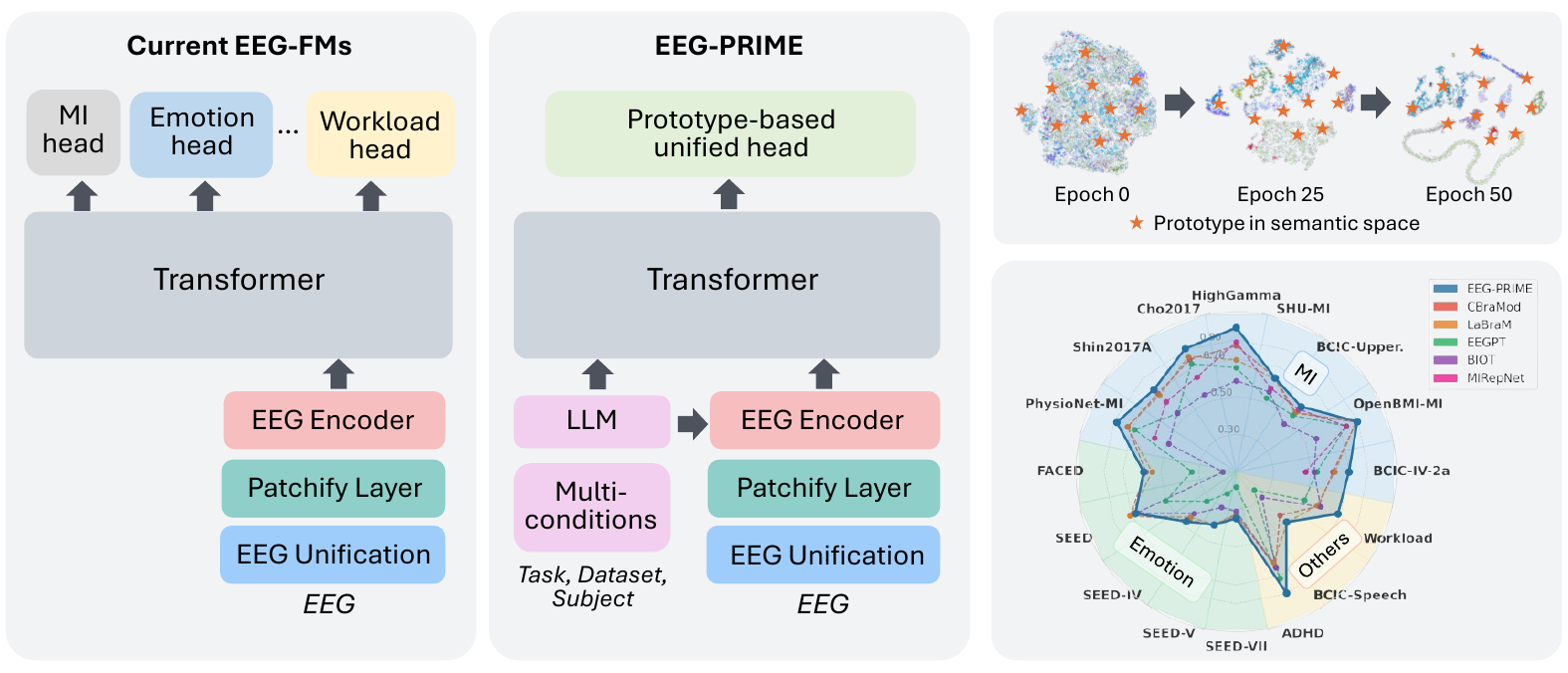}
    \caption{Overview of EEG-PRIME and comparison with existing EEG foundation models.
    \textbf{Left:} Existing EEG-FMs rely on task-specific classifier heads (e.g., separate heads for MI, emotion, and workload), which require retraining for each new task and cannot generalize to unseen label spaces.
    \textbf{Right:} EEG-PRIME introduces multi-level conditioning (task instruction, dataset identity, and subject-invariance constraint) into the EEG encoder via a Q-Former, and replaces all task-specific heads with a single unified prototype-based classifier, where class prototypes are constructed as frozen text embeddings of category label strings.
    \textbf{Top right:} t-SNE visualization of EEG trial embeddings at different training epochs, showing how multi-level conditioning progressively organizes representations into well-separated class clusters.
    \textbf{Bottom right:} Radar chart comparing EEG-PRIME against prior EEG foundation models across all 16 downstream datasets under dataset-specific fine-tuning, demonstrating consistent improvements across diverse BCI paradigms.}
    \label{fig:overview}
\end{figure*}

To address these challenges, we propose \textbf{EEG-PRIME}, an EEG foundation model using multi-level conditioning for cross-dataset multi-task EEG decoding (Fig.~\ref{fig:overview}). Our central hypothesis is that \textbf{language should not directly alter EEG representations; instead, it should guide how the model queries and interprets EEG at multiple semantic levels, thereby enabling more robust and subject-invariant neural decoding}. This stands in contrast to prior approaches that inject language by concatenating text tokens with EEG tokens, which conflates the two modalities and risks distorting the neural signal. Specifically, EEG-PRIME performs instruction tuning driven by  a language-guided tri-level conditioning framework. Task condition specifies the semantic objective of decoding, such as MI, emotion recognition, or mental workload classification. It is encoded by a frozen text encoder and injected into the Q-Former through Layer-wise Query Modulation (LQM), thereby guiding how the model interprets EEG signals for different tasks. Dataset condition captures dataset-specific characteristics that are not fully described by task semantics alone. We represent each dataset with a learnable embedding, which is fused with the task instruction embedding and used to modulate the Q-Former at every layer. Subject condition is used to reduce subject-specific bias and improve cross-subject generalization. Instead of directly modulating the network, it is incorporated through adversarial training with a gradient reversal layer, encouraging the shared representation to remain task-discriminative while being less predictive of subject identity.

Our main contributions are as follows:
\begin{itemize}[noitemsep,topsep=0pt,leftmargin=*,itemsep=2pt]
\item We introduce \textbf{EEG-PRIME}, an EEG foundation model for cross-dataset multi-task EEG decoding based on \textbf{text-anchored prototype classification}. This design enables unified prediction across heterogeneous label spaces, supporting both zero-shot inference without target-domain training and dataset-specific fine-tuning with optional MLP heads.
\item We propose a \textbf{language-guided tri-level conditioning framework} that provides structured priors at three granularities (task-semantic instructions, dataset-level soft embeddings, and subject-invariance constraints), enabling objective-aware, domain-adaptive, and subject-invariant EEG decoding within a single model.
\item We propose \textbf{Layer-wise Query Modulation (LQM)}, which injects the fused instruction embedding into every Q-Former sublayer via independent scale-shift pairs $(\gamma,\beta)$, enabling instruction-aware control over the latent query space at each transformer layer.
\item We evaluate on \textbf{18} datasets spanning five BCI paradigms (MI, emotion recognition, medical healthcare, covert speech, and mental workload), of which sixteen are used for dataset-specific fine-tuning and two are held out for zero-shot evaluation, and show consistent improvements over strong EEG baselines and prior EEG foundation models under cross-subject/session and zero-shot settings.

\end{itemize}

\section{Related Work}
\label{sec:related}

\subsection{Self-supervised Pretraining.}
Self-supervised pretraining (SSL) has emerged as a powerful paradigm in representation learning, reducing the reliance on large amounts of annotated data while leveraging abundant unlabeled signals.
Self-supervised methods design pretext tasks that encourage models to learn meaningful feature representations from the inherent structure of data.
Early successes in natural language processing, such as BERT \citep{devlin2019bert} and the GPT series \citep{radford2019language}, demonstrated that masked language modeling and next-word prediction can yield representations transferable to diverse downstream tasks.
The rise of EEG foundation models is closely tied to the adoption of SSL, which addresses two persistent challenges in EEG modeling: the scarcity of high-quality annotations and the substantial heterogeneity across subjects, devices, montages, and experimental paradigms. SSL has become a central paradigm for EEG representation learning because it enables models to exploit large amounts of unlabeled neural signals and transfer the learned representations to downstream decoding tasks. Recent EEG-specific foundation models have largely adopted masked or generative SSL objectives, with LaBraM~\citep{jiang2024large} being a representative example that scales masked token prediction to over 2,500 hours of heterogeneous EEG data. In this line of work, SSL is not merely used as a regularizer, but as the central mechanism for building transferable neural representations from heterogeneous large-scale data.
Within this masked-pretraining line, several studies have explored how to make the reconstruction task more informative. EEG2Rep \citep{mohammadi2024eeg2rep} argued that conventional random masking may produce suboptimal supervision for EEG, and proposed informative masked inputs through a semantic subsequence preserving strategy. Rather than masking arbitrarily, EEG2Rep preserves informative subsequences and predicts masked signals in latent representation space, thereby increasing the semantic difficulty of the pretext task and encouraging richer EEG representations.
WAVELET2VEC \citep{peng2023wavelet2vec} proposed a filter-bank masked autoencoder for EEG-based seizure subtype classification, where wavelet or filter-bank analysis is used to generate multi-grained time-frequency representations before masked reconstruction. Compared with purely time-domain masking, this approach explicitly leverages spectral structure and suggests that frequency-aware reconstruction can be beneficial for self-supervised EEG pretraining, especially in clinically relevant settings.
\subsection{EEG Foundation Models.}
The concept of foundation models has recently expanded into the EEG domain, aiming to build large-scale pre-trained backbones that generalize across datasets, tasks, and clinical conditions.
BIOT \citep{yang2023biot} explored scalable transformer-based architectures for biomedical signals, while EEGPT \citep{wang2024eegpt} leveraged masked prediction and contrastive pretraining to improve generalization across heterogeneous EEG datasets.
LaBraM \citep{jiang2024large} introduced a unified EEG foundation model for cross-dataset pretraining. It segments EEG into channel patches, learns a vector-quantized neural tokenizer via neural spectrum prediction, and pretrains transformers by masked prediction of discrete neural codes, aiming to learn generic EEG representations across heterogeneous BCI tasks.
CBraMod \citep{wang2024cbramod} further improves EEG foundation modeling by explicitly separating spatial and temporal dependencies. Its criss-cross transformer uses parallel attention mechanisms together with asymmetric conditional positional encoding, enabling better adaptation to EEG recordings with diverse formats while maintaining strong cross-task generalization.
For MI specifically, MIRepNet \citep{liu2025mirepnet} proposed a dedicated MI-oriented foundation model rather than a general-purpose EEG backbone. It combines a neurophysiologically informed preprocessing pipeline and channel template with a hybrid pretraining strategy that integrates masked token reconstruction and supervised MI classification, leading to strong adaptation on downstream MI datasets.
PhysioOmni \citep{jiang2025towards} extends foundation modeling from EEG to multimodal physiological signals. It trains a decoupled multimodal tokenizer over signals such as EEG, ECG, EOG, and EMG to separate modality-invariant and modality-specific information, and further adopts resilient fine-tuning with prototype alignment to remain robust under arbitrary missing modalities.
NeuroLM \citep{jiang2024neurolm} is a universal multi-task EEG foundation model that bridges EEG and language by treating EEG signals as a foreign language for large language models. It learns a text-aligned neural tokenizer to discretize EEG into neural tokens, feeds these tokens into an autoregressive LLM, and then applies multi-task instruction tuning to unify diverse EEG tasks within a single model.

\section{Method}
\label{sec:method}

\begin{figure*}[t]
\graphicspath{{image/}} 
\centerline{\includegraphics[width=2.05\columnwidth]{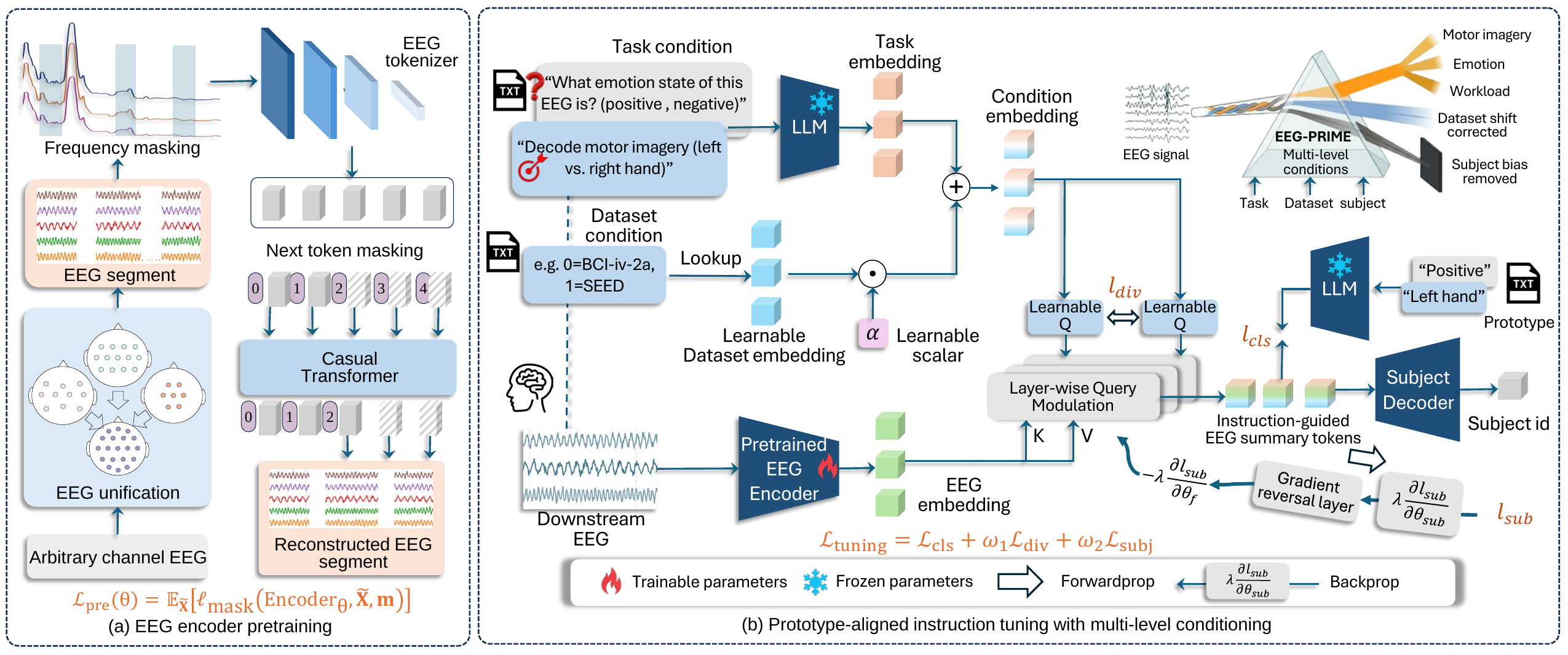}}
\caption{The architecture design of EEG-PRIME. \textbf{(a)} Self-supervised pretraining of an EEG encoder with frequency-cutoff augmentation and masked reconstruction. \textbf{(b)} Prototype-aligned instruction tuning with multi-level conditioning, where task and dataset instruction embeddings modulate intermediate activations via Layer-wise Query Modulation (LQM) and learnable queries summarize EEG tokens through cross-attention.}
\label{fig:LEEGUEMI}
\end{figure*}
In this section, we introduce the design of \textbf{EEG-PRIME}, a language-aligned EEG foundation model for cross-dataset multi-task EEG decoding. EEG-PRIME is trained in two stages: (i) a self-supervised EEG pretraining stage using frequency-cutoff augmentation and masked reconstruction to learn transferable token representations; and (ii) prototype-aligned instruction tuning with multi-level conditioning, where EEG embeddings are conditioned on task instructions, dataset identifiers and aligned with text-defined class prototypes for unified EEG decoding across diverse BCI paradigms. The architecture design of EEG-PRIME can be found in Fig.~\ref{fig:LEEGUEMI}.
\subsection{Problem Formulation}

We study multi-dataset, cross-subject EEG decoding under domain shifts caused by different acquisition protocols, task paradigms, and subject-specific neurophysiology. Let $M$ denote the number of datasets, indexed by $d\in\{1,\ldots,M\}$. Each dataset $d$ provides $N_d$ labeled samples $\{(\mathbf{X}_{d,i}, y_{d,i}, s_{d,i})\}_{i=1}^{N_d}$, where $\mathbf{X}_{d,i}\in\mathbb{R}^{C\times T_d}$ is a multi-channel EEG segment with $C$ channels and dataset-specific length $T_d$, $y_{d,i}\in\mathcal{Y}_d$ is the corresponding class label, and $s_{d,i}\in\{1,\ldots,n_d\}$ is the subject identifier ($n_d$ denotes the number of subjects in dataset $d$). Datasets differ in their label spaces $\mathcal{Y}_d$, precluding a shared classification head across datasets.

Our goal is to learn a model $f_\theta$ that generalizes across datasets and subjects by conditioning on both the dataset identity $d$ and subject identity $s$:
\begin{equation}
\hat{y} \;=\; f_\theta(\mathbf{X}; d, s).
\end{equation}
We train $f_\theta$ by minimizing the expected classification risk over a mixture of dataset distributions:
\begin{equation}
\min_{\theta}\;\;
\mathbb{E}_{d\sim\mathcal{D}}\;
\mathbb{E}_{(\mathbf{X},y,s)\sim \mathcal{S}_d}
\big[\mathcal{L}_{\text{proto}}\big(f_\theta(\mathbf{X}; d, s), y\big)\big],
\end{equation}
where $\mathcal{S}_d$ denotes the data distribution of dataset $d$, and $\mathcal{L}_{\text{proto}}$ is a prototype-based classification loss defined via cosine similarity between the model output and frozen text embeddings of class label strings (detailed in Section~\ref{sec:prototype}). This formulation eliminates dataset-specific classifier heads and enables unified prediction across heterogeneous label spaces.

\subsection{Self-supervised Pretraining of the EEG Encoder}
\label{sec:pretrain}

We first pretrain an EEG encoder to extract task-agnostic token representations from unlabeled EEG segments. 
Given an EEG segment $\mathbf{X}\in\mathbb{R}^{C\times T}$, we align it to a fixed length $L$ via padding/cropping, yielding $\tilde{\mathbf{X}}\in\mathbb{R}^{C\times L}$. 
During training, we use random cropping (when $T>L$) and random left/right zero-padding (when $T<L$) as mild temporal augmentation; validation uses deterministic alignment.
\paragraph{EEG data unification.}
Pretraining data are collected from heterogeneous EEG systems with different channel sets, names and sampling rate. 
To enable unified representation learning, we perform channel and sampling rate unification before feeding it into the EEG encoder. 
Let $\mathcal{C}^{\star}=\{c^{\star}_1,\dots,c^{\star}_{C^\star}\}$ denote the template channels ($C^\star{=}65$), and let a source trial be $\mathbf{X}\in\mathbb{R}^{C\times T}$ with source channel names $\mathcal{C}=\{c_1,\dots,c_C\}$. 
We define a policy-based mapping $\pi:\mathcal{C}\rightarrow \mathcal{C}^{\star}$ that assigns each source channel to one template channel. 
The mapped signal $\mathbf{X}^{\star}\in\mathbb{R}^{C^\star\times T}$ is computed by averaging all source channels that are mapped to the same template channel:
\begin{equation}
\mathbf{X}^{\star}[j,:] \;=\; 
\frac{1}{n_j}\sum_{i:\,\pi(c_i)=c^{\star}_j}\mathbf{X}[i,:],
\qquad
n_j=\Big|\{i:\pi(c_i)=c^{\star}_j\}\Big|.
\end{equation}
If a template channel $c^{\star}_j$ is not observed in the source montage ($n_j=0$), we mark it as missing and interpolate it on the template montage using spherical spline interpolation. 

To standardize the temporal resolution and ensure consistent tokenization and masking behavior, we resample all EEG recordings to a common sampling rate of $f_s=200$~Hz prior to pretraining. 
In the pretrain stage, the EEG is cropped into 10-second segments. This procedure yields a consistent $65$-channel representation and 2000 time points for all datasets and recording configurations, enabling scalable pretraining across heterogeneous sources.

\paragraph{EEG encoder pretraining}
The encoder (denoted by $\mathrm{Encoder}_{\theta}$) maps the aligned EEG to a sequence of latent tokens. Specifically, $\mathrm{Encoder}_{\theta}$ consists of a CNN-based tokenizer that segments $\tilde{\mathbf{X}}$ into non-overlapping temporal windows and extracts per-window feature vectors via depthwise-separable convolutions, followed by a Transformer that models temporal dependencies across tokens:
\begin{equation}
\mathbf{H} = \mathrm{Transformer}\bigl(\mathrm{CNN\text{-}Tokenizer}(\tilde{\mathbf{X}})\bigr)
\in \mathbb{R}^{N \times d_h},
\end{equation}
where $N$ is the number of temporal windows and $d_h$ is the token dimension.

We apply a random token mask with ratio $r$ (set to $r{=}0.5$) and optimize a masked reconstruction objective. 
In addition, we apply a frequency-cutoff augmentation that randomly removes a contiguous frequency band in the Fourier domain, encouraging spectral robustness before tokenization.
Let $\mathbf{m}\in\{0,1\}^{N}$ be a binary mask indicating the masked token positions. 
The pretraining loss can be written as:
\begin{equation}
\mathcal{L}_{\mathrm{pre}}(\theta)
\;=\;
\mathbb{E}_{\tilde{\mathbf{X}}}\Big[
\ell_{\mathrm{mask}}\big(\mathrm{Encoder}_{\theta}, \tilde{\mathbf{X}}, \mathbf{m}\big)
\Big],
\end{equation}
where $\ell_{\mathrm{mask}}(\cdot)$ is the mean squared error between the reconstructed and original EEG signal values at the masked token positions.

\subsection{Prototype-Aligned Instruction Tuning with Multi-level Conditioning}
We propose an instruction-tuned multi-dataset framework for multi-task EEG decoding with three key conditioning components:
(i) instruction-conditioned adaptation via Layer-wise Query Modulation (LQM),
(ii) subject-invariant regularization through adversarial learning, and
(iii) text-anchored prototype classification in a unified semantic label space.

\subsubsection{Instruction-Conditioned Adaptation via LQM}
We introduce instruction-conditioned adaptation to control how the model interprets EEG signals across datasets and task descriptions.
For each dataset $d$, we define a small set of textual instructions $\mathcal{I}_d$ that jointly specify the decoding objective and the candidate class labels (e.g., ``Decode motor imagery (left hand vs.\ right hand)'', ``Recognize the emotion (positive vs.\ negative vs.\ neutral)'', ``Classify the mental workload (low vs.\ high)'').
Each instruction is embedded by the same frozen Sentence-BERT \citep{reimers2019sentence} (SBERT; all-mpnet-base-v2) text encoder, producing $\mathbf{e}_{\text{ins}}\in\mathbb{R}^{D}$.
To capture dataset-specific acquisition differences, we assign each dataset a learnable dataset-level soft embedding $\mathbf{e}_{d}\in\mathbb{R}^{D}$ (initialized to zero) and form a combined conditioning vector
\begin{equation}
\mathbf{e}_{\text{cond}} = \mathbf{e}_{\text{ins}} + \alpha \mathbf{e}_{d},
\end{equation}
where $\alpha$ is a learned scalar scale (initialized to 1 and jointly optimized with the dataset embeddings).

\begin{figure}[t]
\centering
\includegraphics[width=0.95\columnwidth]{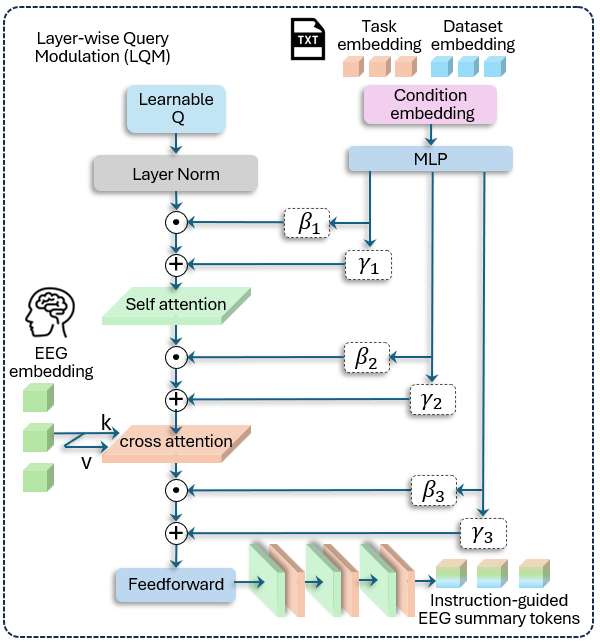}
\caption{Layer-wise Query Modulation (LQM) in the instruction-conditioned Q-Former. The conditioning vector $\mathbf{e}_{\text{cond}}$ generates per-layer scale/shift parameters that modulate the query stream after self-attention, cross-attention, and the feed-forward sublayer.}
\label{fig:adaln_modulation}
\end{figure}

Instead of concatenating text tokens with EEG tokens, we inject the conditioning vector through LQM inside the Q-Former,
making the instruction a conditioning signal that modulates intermediate activations:
\begin{equation}
\text{LQM}(\mathbf{h};\mathbf{e}_{\text{cond}})
=\big(\mathbf{1}+\gamma(\mathbf{e}_{\text{cond}})\big)\odot \text{LN}(\mathbf{h})
+\beta(\mathbf{e}_{\text{cond}}),
\end{equation}
where $\mathbf{h}\in\mathbb{R}^{N_q\times D}$ is the query hidden state after a Q-Former sublayer (self-attention, cross-attention, or feed-forward), $\gamma(\cdot)$ and $\beta(\cdot)$ are learned projections from the conditioning vector, $\odot$ denotes element-wise multiplication, and $\text{LN}(\cdot)$ is layer normalization.
As illustrated in Fig.~\ref{fig:adaln_modulation}, each Q-Former layer produces three $(\gamma,\beta)$ pairs to modulate the query states after self-attention, cross-attention, and the feed-forward sublayer, respectively. This sublayer-level granularity allows the instruction to independently steer query-to-query interactions, EEG-to-query attention, and feature-space transformations, providing finer-grained conditioning than a single per-layer modulation would afford.
For training stability, we initialize the final projection in the LQM parameter generator to zero, ensuring near-identity modulation at the beginning of optimization.

The Q-Former aggregates EEG tokens using a set of learnable query embeddings.
However, without explicit constraints, different query slots may collapse to highly correlated representations, reducing the effective capacity of the latent reader.
To encourage complementary information extraction among query slots, we propose a \emph{query diversity regularization} term that penalizes redundancy across queries throughout the Q-Former stack.
Let $\mathbf{Q}^{(\ell)}\in\mathbb{R}^{N_q\times D}$ denote the query states at the output of the $\ell$-th Q-Former layer after all three sublayers have been applied.
We first $\ell_2$-normalize query vectors along the feature dimension to obtain $\widetilde{\mathbf{Q}}^{(\ell)}$.
We then compute the query-to-query similarity matrix:
\begin{equation}
\mathbf{S}^{(\ell)}=\widetilde{\mathbf{Q}}^{(\ell)} \big(\widetilde{\mathbf{Q}}^{(\ell)}\big)^{\top}\in\mathbb{R}^{N_q\times N_q}.
\end{equation}
To promote diversity, we penalize the off-diagonal similarities (i.e., correlations between different query slots):
\begin{equation}
\mathcal{L}_{\text{div}}
=\frac{1}{L_q}\sum_{\ell=1}^{L_q}
\frac{1}{N_q(N_q-1)}
\sum_{i\neq j}\Big(\mathbf{S}^{(\ell)}_{ij}\Big)^2,
\end{equation}
where $L_q$ is the number of Q-Former layers and $N_q$ is the number of query slots.
This objective encourages different queries to attend to distinct aspects of the EEG token sequence, preventing representational collapse among query slots.
The final Q-Former output is obtained by pooling the $N_q$ query states into a single vector $\mathbf{z}\in\mathbb{R}^D$, which we refer to as the \emph{instruction-guided EEG summary embedding}; it serves as the unified EEG representation for all downstream objectives.

\subsubsection{Subject-Invariant Regularization with Adversarial Training}
To mitigate inter-subject variability while preserving task-discriminative semantics, we apply subject-invariant regularization through adversarial learning.
Specifically, a lightweight subject classifier $g_{\phi}(\cdot)$ predicts the subject label $s$ from the Q-Former output $\mathbf{z}$.
We employ a gradient reversal layer (GRL), which is identity in the forward pass and multiplies gradients by $-\lambda$ in the backward pass:
\begin{equation}
\text{GR}_{\lambda}(\mathbf{z})=\mathbf{z},\qquad
\frac{\partial \text{GR}_{\lambda}}{\partial \mathbf{z}}=-\lambda \mathbf{I}.
\end{equation}
The adversarial subject loss is
\begin{equation}
\mathcal{L}_{\text{subj}}=\text{CE}\big(g_{\phi}(\text{GR}_{\lambda}(\mathbf{z})),\, s\big).
\end{equation}
While $g_{\phi}$ is trained to discriminate subjects, the encoder is encouraged (via GRL) to remove subject-specific cues from $\mathbf{z}$, promoting representations that are predictive of task classes but less predictive of subject identity.

\subsubsection{Text-Anchored Prototype Classification}
\label{sec:prototype}
To enable learning across heterogeneous EEG datasets without dataset-specific classifier heads, we cast EEG decoding as prototype-based matching in a shared semantic space.
Let $\mathcal{D}=\{1,\dots,M\}$ denote the set of datasets and $\mathcal{Y}=\bigcup_{d\in\mathcal{D}}\mathcal{Y}_d$ the union of class names expressed as short texts (e.g., \textit{Left}, \textit{Right}, \textit{Foot}).
We pre-compute a normalized text embedding for each class name using a frozen text encoder, yielding prototypes $\{\mathbf{p}_k\}_{k=1}^{K}$ with $\|\mathbf{p}_k\|_2=1$, and collect them into $\mathbf{P}\in\mathbb{R}^{K\times D}$.
Given an EEG segment $\mathbf{X}$, the EEG--instruction model produces a semantic representation $\mathbf{z}\in\mathbb{R}^{D}$.
Classification is performed by prototype similarity:
\begin{equation}
\label{eq:logits}
\mathbf{o}=\mathbf{z}\mathbf{P}^{\top}\in\mathbb{R}^{K},
\end{equation}
where $\mathbf{o}$ are the logits. We optimize a prototype cross-entropy objective:
\begin{equation}
\mathcal{L}_{\text{cls}}
=-\log \frac{\exp(\mathbf{z}\cdot \mathbf{p}_{y})}{\sum_{k=1}^{K}\exp(\mathbf{z}\cdot \mathbf{p}_{k})}.
\end{equation}
In practice, $\mathbf{z}$ is $\ell_2$-normalized before computing similarities so that $\mathbf{z}\cdot\mathbf{p}_k = \cos(\mathbf{z}, \mathbf{p}_k)$, making the dot-product logits in Eq.~\eqref{eq:logits} equivalent to cosine similarities used at inference. This formulation aligns EEG representations with text-defined class anchors and reduces reliance on dataset-dependent classifier parameters.

The overall instruction-tuning objective combines all three losses:
\begin{equation}
\mathcal{L}_{\text{tuning}}=\mathcal{L}_{\text{cls}}+\omega_1 \mathcal{L}_{\text{div}}+\omega_2 \mathcal{L}_{\text{subj}},
\end{equation}
where $\omega_1$ and $\omega_2$ are loss weights.

\subsection{Inference Procedure}
To evaluate the proposed EEG-PRIME, we consider three inference regimes:
(1)~\textbf{Task-specific fine-tuning}: following multi-task instruction tuning, the model is further fine-tuned on the training split of each target dataset. A lightweight MLP classification head is added on top of the Q-Former output $\mathbf{z}$ and trained jointly with the shared EEG--language parameters to adapt to the target domain's distribution.
(2)~\textbf{Zero-shot inference}: after multi-task instruction tuning, the model is kept fully frozen and directly evaluated on previously unseen target datasets. No additional optimization, calibration, or linear probing is performed on the target data.
(3)~\textbf{In-domain direct inference}: the model remains fully frozen during evaluation. No dataset-specific fine-tuning, classifier training, or adaptation is performed on the target dataset, but the test set belongs to the target dataset seen during multi-task instruction tuning. By eliminating the effects of dataset-specific adaptation and distribution shift, this protocol allows us to systematically investigate the effects of instruction design, dataset-level conditioning, and modulation parameters (e.g., scale and bias), providing a mechanistic understanding of how the model encodes and utilizes EEG information.
For regimes~(2) and~(3), the model is frozen and the final prediction is obtained by
\begin{equation}
\hat{y} = \arg\max_y \cos(\tilde{\mathbf{z}}, \mathbf{p}_y),
\end{equation}
where $\tilde{\mathbf{z}}$ is the $\ell_2$-normalized Q-Former output and $\mathbf{p}_y$ is the text prototype of the corresponding label $y$ (e.g., ``left hand'', ``right hand'').

\section{Experiments}

\subsection{Dataset}

\paragraph{Pretraining dataset}
We use 9 datasets, namely Stieger2021 \citep{stieger2021mindfulness}, SEED-FRA \citep{liu2022identifying}, SEED-GER \citep{liu2022identifying}, SEED-SD \citep{li2025investigating}, SEED-Neg, ChineseEEG \citep{mou2024chineseeeg}, Chisco \citep{zhang2024chisco}, LargeSpanish\citep{valle2024identification}, ThinkOutLoud \citep{nieto2022thinking} as the pretraining datasets. The total duration of these datasets is around 1153 hours.

\paragraph{Downstream Dataset}
We systematically evaluate EEG-PRIME across 18 datasets spanning five BCI paradigms.
\textit{MI:} OpenBMI-MI \citep{lee2019eeg}, BCIC-IV-2a \citep{tangermann2012review}, BCIC-Upperlimb \citep{jeong20222020}, SHU-MI \citep{ma2022large}, HighGamma \citep{schirrmeister2017deep}, Cho2017 \citep{cho2017eeg}, Shin2017A \citep{shin2016open}, PhysioNet-MI \citep{schalk2004bci2000}, Dreyer2023A \citep{pillette2021experimenters}, Weibo2014 \citep{yi2014evaluation}.
\textit{Emotion Recognition:} FACED \citep{chen2023large}, SEED \citep{duan2013differential}, SEED-IV, SEED-V \citep{liu2021comparing}, and SEED-VII \citep{jiang2024seed}.
\textit{Medical Healthcare:} ADHD-AliMotie \citep{AliMotie2020ADHD}.
\textit{Mental Workload:} Mental Workload \citep{zyma2019electroencephalograms}.
\textit{Covert Speech:} BCIC-Speech. We report the train/validation/test subject splits for all downstream datasets in Table~\ref{tab:dataset_splits}. All datasets use a cross-subject protocol, with 20\% of the training subjects further held out for validation. To evaluate our method in the zero-shot setting, two datasets are held out from all training and fine-tuning stages: Dreyer2023A and Weibo2014, neither of which is seen during instruction tuning or adaptation.

\subsection{Experimental Setup}

\paragraph{Baselines \& Metrics} 
In this paper, we selected both the state-of-the-art traditional models and the EEG-FMs as baselines. For the traditional models, we selected EEGNet \citep{lawhern2018eegnet}, TSception \citep{ding2022tsception}, ST-Transformer \citep{song2021transformer} and Conformer \citep{song2022eeg}. For the EEG foundation model, we selected BIOT \citep{yang2023biot}, EEGPT \citep{wang2024eegpt}, LaBraM \citep{jiang2024large}, CBraMod \citep{wang2024cbramod}. For MI foundation model we selected MIRepNet \citep{liu2025mirepnet} as an additional baseline. To provide a reliable evaluation across imbalanced datasets, we adopted balanced accuracy and Cohen’s Kappa as performance metrics. Balanced accuracy accounts for class imbalance by averaging recall across classes, while Cohen’s Kappa measures the agreement between predicted and true labels beyond chance level, providing a more robust assessment of model performance.

\paragraph{Baseline Implementation Details}
To ensure a fair comparison, all baseline models were re-trained or fine-tuned using their officially released implementations and recommended hyperparameters. For each dataset, EEG trials were resampled to 200 Hz, truncated to a fixed length of 800 samples (65 channels). For transformer-based foundation models, further segmented into non-overlapping 200-sample windows as tokens. Unless otherwise specified, we trained baselines for 100 epochs. We evaluated balanced accuracy, and Cohen’s Kappa on the validation and test splits. Pretrained checkpoints of LaBraM, EEGPT, and CBraMod were loaded and then fine-tuned end-to-end from the official checkpoints. For MIRepNet, we loaded the official pretrained checkpoint and fine-tuned up to 50 epochs with early stopping (patience=10).

\subsection{EEG Preprocessing}
\begin{figure}[h]
\graphicspath{{image/}}
\centerline{\includegraphics[width=1\columnwidth]{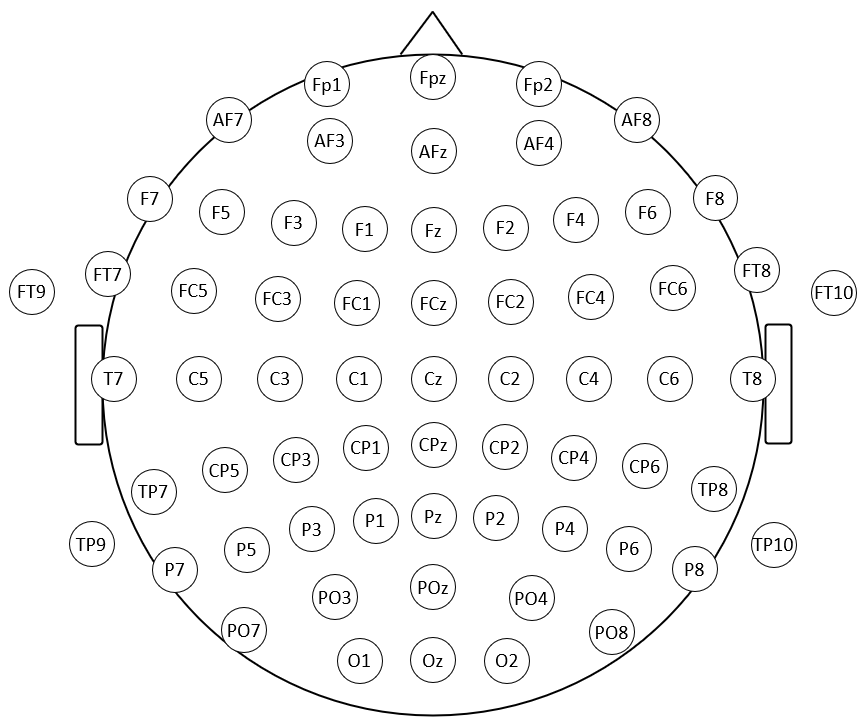}}
\caption{EEG-PRIME employs the 10--10 system with 65 EEG electrodes; any input montage is interpolated to this configuration before being fed into model.}
\label{LEEGUEMI_Montage}
\end{figure}

EEG recordings from different studies typically use diverse electrode montages. EEG-PRIME performs channel unification by interpolating all signals onto the standardized 10--10 electrode layout with 65 channels, as illustrated in Fig.~\ref{LEEGUEMI_Montage}. For datasets recorded with fewer than 65 channels, we perform spatial interpolation to enforce a consistent topological structure across inputs. We downsample all recordings to 200 Hz and apply paradigm-appropriate band-pass filters: 0.3--40 Hz for MI datasets and 0.3--70 Hz for emotion recognition, ADHD, covert speech, and mental workload datasets.

\begin{table}[h]
\centering
\caption{Train/validation/test splits for downstream EEG datasets. CS = Cross-Subject; CT = Cross-Trial; ZS = Zero-Shot (no training or adaptation on this dataset).}
\label{tab:dataset_splits}
\footnotesize
\renewcommand{\arraystretch}{1.1}
\begin{tabular}{l c l l}
\toprule
\textbf{Dataset} & \textbf{Split} & \textbf{Train/Val} & \textbf{Test} \\
\midrule
\multicolumn{4}{@{}l}{\textit{MI}} \\
BCIC-IV-2a     & CS & Subj.\ 1--7   & Subj.\ 8--9 \\
OpenBMI-MI     & CS & Subj.\ 1--42  & Subj.\ 43--54 \\
BCIC-Upperlimb & CS & Subj.\ 1--11  & Subj.\ 12--15 \\
SHU-MI         & CS & Subj.\ 1--20  & Subj.\ 21--25 \\
HighGamma      & CS & Subj.\ 1--10  & Subj.\ 11--14 \\
Cho2017        & CS & Subj.\ 1--40  & Subj.\ 41--49 \\
Shin2017A      & CS & Subj.\ 1--22  & Subj.\ 23--28 \\
PhysioNet-MI   & CS & Subj.\ 1--80  & Subj.\ 81--109 \\
Dreyer2023A    & ZS & ---           & All 60 Subj. \\
Weibo2014      & ZS & ---           & 9 of 10 Subj.$^{*}$ \\
\midrule
\multicolumn{4}{@{}l}{\textit{Emotion Recognition}} \\
FACED          & CS & Subj.\ 1--100 & Subj.\ 101--122 \\
SEED           & CT & Trial 1--9    & Trial 10--12 \\
SEED-IV        & CT & Trial 1--16   & Trial 17--24 \\
SEED-V         & CT & Trial 1--5    & Trial 6--10 \\
SEED-VII       & CT & Trial 1--10   & Trial 11--15 \\
\midrule
\multicolumn{4}{@{}l}{\textit{Covert Speech / ADHD / Mental Workload}} \\
ADHD-AliMotie  & CS & Subj.\ 1--80  & Subj.\ 81--120 \\
BCIC-Speech    & CT & Trial 1--250  & Trial 301--350 \\
Mental Workload & CS & Subj.\ 0--31 & Subj.\ 32--35 \\
\bottomrule
\multicolumn{4}{@{}l}{\footnotesize $^{*}$One subject excluded as a known BCI-illiterate participant.}
\end{tabular}
\end{table}

\paragraph{Implementation Details}
Training proceeds in two sequential stages, with all parameters updated jointly within each stage.
\textbf{Stage~1 (pretraining):} The EEG encoder (tokenizer: 0.27\,M; transformer: 9.98\,M) is pretrained on unlabeled EEG with AdamW using two learning-rate groups: $1\times10^{-3}$ for the tokenizer and $1\times10^{-4}$ for the remaining encoder parameters, weight decay $1\times10^{-3}$, cosine schedule with 5 warm-up epochs, for 100 epochs total, batch size 256, gradient clipping at 1.0.
\textbf{Stage~2 (instruction tuning):} The pretrained EEG encoder is frozen and the Q-Former (4 layers, 8 queries; 10.53\,M), dataset embeddings (0.01\,M), and LQM projections are trained with AdamW ($\text{lr}=1\times10^{-3}$, weight decay $1\times10^{-3}$), cosine schedule with 5 warm-up epochs, 100 epochs, batch size 256. The adversarial subject-invariance loss is linearly ramped in over 8 epochs with weight $\omega_2{=}0.5$ and GRL scale $\lambda{=}0.03$; the diversity loss weight is $\omega_1{=}1\times10^{-3}$. Total trainable parameters: 20.79\,M. All experiments were conducted on 2×NVIDIA RTX PRO 5000 GPUs.
\section{Result}

\subsection{Comparison with State-of-the-Art}
\label{sec:main_results}

\begin{table*}[t]
\centering
\captionsetup[subtable]{font=footnotesize, skip=1pt}
\caption{Performance comparison between EEG-PRIME (ours) and baselines on 16 datasets, reported as mean~$\pm$~std over three random seeds.}
\label{tab:mean_std}
\renewcommand{\arraystretch}{1.0}

\begin{subtable}{0.9\textwidth}
\centering
\caption{Motor Imagery (Part I)}
\scriptsize
\begin{adjustbox}{width=\textwidth}
\begin{tabular}{l cc cc cc cc}
\toprule
\multirow{2}{*}{Methods}
& \multicolumn{2}{c}{BCIC-IV-2a}
& \multicolumn{2}{c}{OpenBMI-MI}
& \multicolumn{2}{c}{BCIC-Upperlimb}
& \multicolumn{2}{c}{SHU-MI} \\
\cmidrule(lr){2-3}\cmidrule(lr){4-5}\cmidrule(lr){6-7}\cmidrule(lr){8-9}
& B.Acc & Kappa & B.Acc & Kappa & B.Acc & Kappa & B.Acc & Kappa \\
\midrule
EEGNet    & \cellcolor{green!20}0.6369$\pm$0.0128 & \cellcolor{green!20}0.4685$\pm$0.0171 & \cellcolor{green!20}0.8170$\pm$0.0216 & \cellcolor{green!20}0.6259$\pm$0.0432 & 0.5281$\pm$0.0113 & 0.2962$\pm$0.0170 & 0.5665$\pm$0.0060 & 0.1976$\pm$0.0120 \\
Conformer & \cellcolor{green!10}0.6347$\pm$0.0110 & \cellcolor{green!10}0.4663$\pm$0.0147 & \cellcolor{green!35}\textbf{0.8274$\pm$0.0025} & \cellcolor{green!35}\textbf{0.6387$\pm$0.0050} & \cellcolor{green!10}0.5473$\pm$0.0181 & 0.3120$\pm$0.0271 & 0.5780$\pm$0.0147 & 0.1591$\pm$0.0295 \\
TScep.    & 0.6252$\pm$0.0050 & 0.4543$\pm$0.0067 & 0.6665$\pm$0.0032 & 0.3425$\pm$0.0064 & 0.5199$\pm$0.0115 & 0.2792$\pm$0.0172 & 0.5564$\pm$0.0093 & 0.2047$\pm$0.0185 \\
STTran.   & 0.5816$\pm$0.0175 & 0.4334$\pm$0.0234 & 0.7514$\pm$0.0026 & 0.5046$\pm$0.0051 & 0.5284$\pm$0.0243 & 0.2972$\pm$0.0364 & 0.5714$\pm$0.0077 & 0.2191$\pm$0.0154 \\
BIOT      & 0.5139$\pm$0.0274 & 0.3898$\pm$0.0366 & 0.5613$\pm$0.0023 & 0.1225$\pm$0.0047 & 0.4595$\pm$0.0651 & 0.1843$\pm$0.0976 & 0.5587$\pm$0.0353 & 0.2151$\pm$0.0706 \\
EEGPT     & 0.5279$\pm$0.0166 & 0.4617$\pm$0.0221 & 0.7323$\pm$0.0016 & 0.4624$\pm$0.0031 & 0.5231$\pm$0.0143 & 0.2031$\pm$0.0214 & 0.5228$\pm$0.0158 & 0.1657$\pm$0.0316 \\
LaBraM    & 0.6234$\pm$0.0351 & 0.4599$\pm$0.0468 & 0.7874$\pm$0.0033 & 0.5765$\pm$0.0065 & 0.5414$\pm$0.0258 & 0.3108$\pm$0.0386 & \cellcolor{green!10}0.6338$\pm$0.0510 & \cellcolor{green!10}0.2338$\pm$0.1020 \\
CBraMod   & 0.6139$\pm$0.0299 & 0.4385$\pm$0.0398 & 0.7895$\pm$0.0855 & 0.5828$\pm$0.1709 & 0.5454$\pm$0.0768 & \cellcolor{green!10}0.3145$\pm$0.1152 & \cellcolor{green!35}\textbf{0.6403$\pm$0.0107} & \cellcolor{green!20}0.2384$\pm$0.0214 \\
MIRepNet  & 0.4686$\pm$0.0230 & 0.2914$\pm$0.0306 & 0.7327$\pm$0.0077 & 0.4654$\pm$0.0153 & \cellcolor{green!20}0.5683$\pm$0.0228 & \cellcolor{green!20}0.3529$\pm$0.0342 & 0.5798$\pm$0.0148 & 0.1595$\pm$0.0298 \\
\midrule
\rowcolor{blue!8}
EEG-PRIME      & \cellcolor{green!35}\textbf{0.6982$\pm$0.0124} & \cellcolor{green!35}\textbf{0.5975$\pm$0.0165} & \cellcolor{green!10}0.7956$\pm$0.0055 & \cellcolor{green!10}0.5911$\pm$0.0110 & \cellcolor{green!35}\textbf{0.5879$\pm$0.0170} & \cellcolor{green!35}\textbf{0.3837$\pm$0.0252} & \cellcolor{green!20}0.6387$\pm$0.0083 & \cellcolor{green!35}\textbf{0.2774$\pm$0.0167} \\
\bottomrule
\end{tabular}
\end{adjustbox}
\end{subtable}
\vspace{2pt}

\begin{subtable}{0.9\textwidth}
\centering
\caption{Motor Imagery (Part II)}
\scriptsize
\begin{adjustbox}{width=\textwidth}
\begin{tabular}{l cc cc cc cc}
\toprule
\multirow{2}{*}{Methods}
& \multicolumn{2}{c}{HighGamma}
& \multicolumn{2}{c}{Cho2017}
& \multicolumn{2}{c}{Shin2017A}
& \multicolumn{2}{c}{PhysioNet-MI} \\
\cmidrule(lr){2-3}\cmidrule(lr){4-5}\cmidrule(lr){6-7}\cmidrule(lr){8-9}
& B.Acc & Kappa & B.Acc & Kappa & B.Acc & Kappa & B.Acc & Kappa \\
\midrule
EEGNet    & \cellcolor{green!10}0.7856$\pm$0.0158 & \cellcolor{green!10}0.5838$\pm$0.0238 & \cellcolor{green!10}0.7644$\pm$0.0047 & \cellcolor{green!10}0.5310$\pm$0.0094 & \cellcolor{green!20}0.7054$\pm$0.0137 & \cellcolor{green!35}0.4521$\pm$0.0273 & 0.6953$\pm$0.0076 & 0.3980$\pm$0.0153 \\
Conformer & 0.7637$\pm$0.0318 & 0.5627$\pm$0.0477 & \cellcolor{green!20}0.7838$\pm$0.0057 & \cellcolor{green!20}0.5719$\pm$0.0115 & 0.6464$\pm$0.0080 & 0.2901$\pm$0.0159 & 0.6951$\pm$0.0217 & 0.3952$\pm$0.0435 \\
TScep.    & 0.6851$\pm$0.0210 & 0.5129$\pm$0.0315 & 0.7339$\pm$0.0055 & 0.4640$\pm$0.0110 & 0.5992$\pm$0.0176 & 0.1891$\pm$0.0352 & 0.6649$\pm$0.0119 & 0.3216$\pm$0.0238 \\
STTran.   & 0.7019$\pm$0.0627 & 0.5162$\pm$0.0941 & 0.7612$\pm$0.0147 & 0.5260$\pm$0.0295 & 0.6168$\pm$0.0171 & 0.2371$\pm$0.0342 & 0.6706$\pm$0.0097 & 0.3370$\pm$0.0196 \\
BIOT      & 0.5824$\pm$0.0090 & 0.4855$\pm$0.0135 & 0.5413$\pm$0.0159 & 0.0729$\pm$0.0318 & 0.5422$\pm$0.0052 & 0.0909$\pm$0.0105 & 0.4874$\pm$0.0061 & 0.0162$\pm$0.0122 \\
EEGPT     & 0.6516$\pm$0.0252 & 0.5152$\pm$0.0378 & 0.7197$\pm$0.0029 & 0.4305$\pm$0.0058 & 0.5389$\pm$0.0297 & 0.0688$\pm$0.0595 & 0.6820$\pm$0.0140 & 0.3730$\pm$0.0279 \\
LaBraM    & 0.6939$\pm$0.0078 & 0.5200$\pm$0.0117 & 0.7614$\pm$0.1215 & 0.5221$\pm$0.2430 & 0.6744$\pm$0.0086 & 0.3607$\pm$0.0172 & \cellcolor{green!20}0.7246$\pm$0.0069 & \cellcolor{green!20}0.4487$\pm$0.0138 \\
CBraMod   & 0.7684$\pm$0.0101 & 0.5671$\pm$0.0151 & 0.7439$\pm$0.0977 & 0.4974$\pm$0.1954 & \cellcolor{green!10}0.6861$\pm$0.0047 & \cellcolor{green!10}0.3689$\pm$0.0094 & \cellcolor{green!10}0.7219$\pm$0.0920 & \cellcolor{green!10}0.4386$\pm$0.1842 \\
MIRepNet  & \cellcolor{green!20}0.7886$\pm$0.0058 & \cellcolor{green!20}0.6829$\pm$0.0087 & 0.6433$\pm$0.0055 & 0.2867$\pm$0.0110 & 0.6246$\pm$0.0091 & 0.2496$\pm$0.0181 & 0.5684$\pm$0.0023 & 0.1367$\pm$0.0045 \\
\midrule
\rowcolor{blue!8}
EEG-PRIME      & \cellcolor{green!35}\textbf{0.8639$\pm$0.0041} & \cellcolor{green!35}\textbf{0.7959$\pm$0.0062} & \cellcolor{green!35}\textbf{0.8065$\pm$0.0042} & \cellcolor{green!35}\textbf{0.6130$\pm$0.0083} & \cellcolor{green!35}\textbf{0.7168$\pm$0.0190} & \cellcolor{green!20}0.4328$\pm$0.0380 & \cellcolor{green!35}\textbf{0.7852$\pm$0.0048} & \cellcolor{green!35}\textbf{0.5703$\pm$0.0096} \\
\bottomrule
\end{tabular}
\end{adjustbox}
\end{subtable}

\vspace{2pt}

\begin{subtable}{0.9\textwidth}
\centering
\caption{Emotion Recognition}
\scriptsize
\begin{adjustbox}{width=\textwidth}
\begin{tabular}{l cc cc cc cc}
\toprule
\multirow{2}{*}{Methods}
& \multicolumn{2}{c}{FACED}
& \multicolumn{2}{c}{SEED}
& \multicolumn{2}{c}{SEED-IV}
& \multicolumn{2}{c}{SEED-V} \\
\cmidrule(lr){2-3}\cmidrule(lr){4-5}\cmidrule(lr){6-7}\cmidrule(lr){8-9}
& B.Acc & Kappa & B.Acc & Kappa & B.Acc & Kappa & B.Acc & Kappa \\
\midrule
EEGNet    & 0.4271$\pm$0.0029 & 0.3512$\pm$0.0032 & 0.5337$\pm$0.0018 & 0.3143$\pm$0.0027 & 0.3651$\pm$0.0152 & 0.1578$\pm$0.0173 & 0.2932$\pm$0.0117 & 0.1136$\pm$0.0162 \\
Conformer & 0.4943$\pm$0.0034 & 0.4263$\pm$0.0044 & 0.6254$\pm$0.0075 & 0.4342$\pm$0.0111 & 0.4094$\pm$0.0067 & 0.2121$\pm$0.0121 & 0.3060$\pm$0.0190 & 0.1335$\pm$0.0239 \\
TScep.    & 0.2056$\pm$0.0076 & 0.1088$\pm$0.0092 & 0.6369$\pm$0.0072 & 0.4604$\pm$0.0108 & 0.4063$\pm$0.0097 & 0.1876$\pm$0.0142 & 0.3637$\pm$0.0044 & 0.1984$\pm$0.0052 \\
STTran.   & 0.3791$\pm$0.0113 & 0.2999$\pm$0.0123 & 0.5882$\pm$0.0098 & 0.3873$\pm$0.0145 & 0.3616$\pm$0.0035 & 0.1415$\pm$0.0040 & 0.2244$\pm$0.0058 & 0.0344$\pm$0.0052 \\
BIOT      & 0.1711$\pm$0.0123 & 0.0647$\pm$0.0131 & 0.6674$\pm$0.0051 & 0.5034$\pm$0.0076 & 0.4141$\pm$0.0069 & 0.1937$\pm$0.0048 & 0.3045$\pm$0.0105 & 0.1306$\pm$0.0113 \\
EEGPT     & 0.3346$\pm$0.0014 & 0.2486$\pm$0.0022 & 0.5054$\pm$0.0082 & 0.2659$\pm$0.0124 & 0.3202$\pm$0.0069 & 0.0861$\pm$0.0071 & 0.2253$\pm$0.0057 & 0.0335$\pm$0.0064 \\
LaBraM    & \cellcolor{green!10}0.5457$\pm$0.0150 & \cellcolor{green!10}0.4809$\pm$0.0160 & \cellcolor{green!20}0.7083$\pm$0.0018 & \cellcolor{green!20}0.5613$\pm$0.0027 & \cellcolor{green!10}0.4415$\pm$0.0010 & \cellcolor{green!10}0.2560$\pm$0.0027 & \cellcolor{green!10}0.4010$\pm$0.0016 & \cellcolor{green!10}0.2563$\pm$0.0009 \\
CBraMod   & \cellcolor{green!20}0.5787$\pm$0.0123 & \cellcolor{green!20}0.4941$\pm$0.0120 & \cellcolor{green!35}\textbf{0.7102$\pm$0.0065} & \cellcolor{green!35}\textbf{0.5868$\pm$0.0097} & \cellcolor{green!20}0.4605$\pm$0.0109 & \cellcolor{green!20}0.2728$\pm$0.0167 & \cellcolor{green!20}0.4029$\pm$0.0091 & \cellcolor{green!20}0.2570$\pm$0.0114 \\
\midrule
\rowcolor{blue!8}
EEG-PRIME      & \cellcolor{green!35}\textbf{0.5908$\pm$0.0029} & \cellcolor{green!35}\textbf{0.5340$\pm$0.0035} & \cellcolor{green!10}0.6782$\pm$0.0082 & \cellcolor{green!10}0.5203$\pm$0.0125 & \cellcolor{green!35}\textbf{0.4728$\pm$0.0128} & \cellcolor{green!35}\textbf{0.2804$\pm$0.0025} & \cellcolor{green!35}\textbf{0.4051$\pm$0.0037} & \cellcolor{green!35}\textbf{0.2586$\pm$0.0038} \\
\bottomrule
\end{tabular}
\end{adjustbox}
\end{subtable}

\vspace{2pt}

\begin{subtable}{0.9\textwidth}
\centering
\caption{Emotion (SEED-VII), ADHD, Covert Speech, and Workload}
\scriptsize
\begin{adjustbox}{width=\textwidth}
\begin{tabular}{l cc cc cc cc}
\toprule
\multirow{2}{*}{Methods}
& \multicolumn{2}{c}{SEED-VII}
& \multicolumn{2}{c}{ADHD}
& \multicolumn{2}{c}{BCIC-Speech}
& \multicolumn{2}{c}{Mental Workload} \\
\cmidrule(lr){2-3}\cmidrule(lr){4-5}\cmidrule(lr){6-7}\cmidrule(lr){8-9}
& B.Acc & Kappa & B.Acc & Kappa & B.Acc & Kappa & B.Acc & Kappa \\
\midrule
EEGNet    & 0.2587$\pm$0.0073 & 0.1413$\pm$0.0068 & 0.6349$\pm$0.0412 & 0.2789$\pm$0.0812 & 0.2699$\pm$0.0159 & 0.0880$\pm$0.0198 & 0.5480$\pm$0.0619 & 0.0982$\pm$0.1331 \\
Conformer & 0.3209$\pm$0.0178 & 0.2081$\pm$0.0193 & 0.7316$\pm$0.0329 & 0.4681$\pm$0.0723 & 0.4170$\pm$0.0112 & 0.2722$\pm$0.0140 & 0.5984$\pm$0.0026 & 0.1801$\pm$0.0179 \\
TScep.    & \cellcolor{green!10}0.3300$\pm$0.0049 & \cellcolor{green!10}0.2198$\pm$0.0025 & \cellcolor{green!10}0.7333$\pm$0.0257 & \cellcolor{green!10}0.4752$\pm$0.0552 & \cellcolor{green!35}\textbf{0.5314$\pm$0.0139} & \cellcolor{green!35}\textbf{0.4164$\pm$0.0173} & \cellcolor{green!20}0.6480$\pm$0.0092 & \cellcolor{green!20}0.3034$\pm$0.0235 \\
STTran.   & 0.1867$\pm$0.0051 & 0.0516$\pm$0.0038 & \cellcolor{green!20}0.7515$\pm$0.0067 & \cellcolor{green!20}0.5047$\pm$0.0123 & 0.4266$\pm$0.0103 & 0.2832$\pm$0.0129 & \cellcolor{green!10}0.6335$\pm$0.0786 & \cellcolor{green!10}0.2683$\pm$0.1559 \\
BIOT      & 0.3096$\pm$0.0081 & 0.1939$\pm$0.0077 & 0.6516$\pm$0.0103 & 0.3155$\pm$0.0236 & 0.2920$\pm$0.0569 & 0.1138$\pm$0.0711 & 0.5857$\pm$0.0265 & 0.1480$\pm$0.0228 \\
EEGPT     & 0.1809$\pm$0.0081 & 0.0476$\pm$0.0072 & 0.7124$\pm$0.0459 & 0.4217$\pm$0.1016 & 0.2364$\pm$0.0072 & 0.0478$\pm$0.0091 & 0.4925$\pm$0.0126 & 0.0223$\pm$0.0243 \\
LaBraM    & 0.3244$\pm$0.0040 & 0.2159$\pm$0.0038 & 0.6194$\pm$0.0079 & 0.2285$\pm$0.0147 & \cellcolor{green!20}0.4819$\pm$0.0062 & \cellcolor{green!20}0.3863$\pm$0.0077 & 0.5650$\pm$0.0373 & 0.1643$\pm$0.0636 \\
CBraMod   & \cellcolor{green!20}0.3311$\pm$0.0070 & \cellcolor{green!20}0.2233$\pm$0.0067 & 0.6434$\pm$0.0066 & 0.3121$\pm$0.0186 & 0.4280$\pm$0.0049 & 0.2860$\pm$0.0061 & 0.5746$\pm$0.0279 & 0.1695$\pm$0.0473 \\
\midrule
\rowcolor{blue!8}
EEG-PRIME      & \cellcolor{green!35}\textbf{0.3464$\pm$0.0024} & \cellcolor{green!35}\textbf{0.2395$\pm$0.0033} & \cellcolor{green!35}\textbf{0.7968$\pm$0.0523} & \cellcolor{green!35}\textbf{0.5890$\pm$0.1031} & \cellcolor{green!10}0.4769$\pm$0.0267 & \cellcolor{green!10}0.3461$\pm$0.0334 & \cellcolor{green!35}\textbf{0.6843$\pm$0.0032} & \cellcolor{green!35}\textbf{0.4326$\pm$0.0233} \\
\bottomrule
\end{tabular}
\end{adjustbox}
\end{subtable}

\end{table*}

Table~\ref{tab:mean_std} reports task-specific fine-tuning performance across sixteen datasets spanning five BCI paradigms.
Overall, EEG-PRIME achieves the best balanced accuracy or Kappa on 13 out of 16 datasets and ranks in the top-3 on all datasets, demonstrating consistent cross-paradigm generalization that no single baseline achieves.

\paragraph{MI}
EEG-PRIME achieves the best B.Acc or Kappa on seven out of eight MI datasets, including BCI-IV-2a, BCIC-Upperlimb, SHU-MI, HighGamma, Cho2017, Shin2017A, and PhysioNet-MI. 
On the OpenBMI-MI dataset, EEG-PRIME attains a balanced accuracy of 0.7956, ranking behind Conformer (0.8274) and EEGNet (0.8170).
To provide a comprehensive comparison, we compute the unweighted average performance across all eight datasets. EEG-PRIME achieves the best overall results (B.Acc: 0.7366, Kappa: 0.5327), outperforming CBraMod (B.Acc: 0.6887, Kappa: 0.4308) and EEGNet (B.Acc: 0.6874, Kappa: 0.4441).
These results demonstrate that EEG-PRIME not only improves classification accuracy, but also yields more reliable agreement across diverse MI paradigms.

\paragraph{Emotion Recognition}
Emotion decoding is notably harder, reflected in lower absolute B.Acc values. EEG-PRIME leads on four of the five emotion datasets, namely FACED (0.5908), SEED-IV (0.4728), SEED-V (0.4051), and SEED-VII (0.3464). It also achieves the best kappa on these four datasets, i.e., FACED (0.5340), SEED-IV (0.2804), SEED-V (0.2586), and SEED-VII (0.2395). On the SEED dataset, EEG-PRIME reaches a B.Acc of 0.6782 and a Kappa of 0.5203, ranking behind CBraMod (B.Acc: 0.7102, Kappa: 0.5868) and LaBraM (B.Acc: 0.7083, Kappa: 0.5613). The unweighted average performance shows EEG-PRIME achieves the best overall average balanced accuracy (0.4987), slightly ahead of CBraMod (0.4967) and LaBraM (0.4842).

\paragraph{ADHD, Covert Speech, and Mental Workload}

On the ADHD dataset, EEG-PRIME achieves the highest balanced accuracy (0.7968). On the selected mental workload dataset, EEG-PRIME achieves the top result (0.6843). On the BCIC-Speech dataset, EEG-PRIME reaches a B.Acc of 0.4769 and a Kappa of 0.3461, ranking behind TSception (B.Acc: 0.5314, Kappa: 0.4164) and LaBraM (B.Acc: 0.4819, Kappa: 0.3863). These results show the EEG-PRIME's strong robustness across clinically and cognitively distinct EEG decoding tasks.

To assess whether the performance improvements of EEG-PRIME are statistically meaningful, we conduct pairwise one-sided Wilcoxon signed-rank tests between EEG-PRIME and each baseline, treating each benchmark dataset as an independent observation. Effect sizes are quantified using Cohen's $d$ on the per-dataset balanced accuracy differences. As shown in Table~\ref{tab:stat}, EEG-PRIME significantly outperforms all nine baselines ($p \leq 0.004$), with large effect sizes throughout (Cohen's $d = 0.86$--$2.53$). Against the two strongest competitors, LaBraM and CBraMod, EEG-PRIME wins on 14 out of 16 datasets ($d = 0.86$ and $d = 0.91$, respectively); the two losses for each are isolated to specific datasets (SEED and BCIC-Speech for LaBraM; SHU-MI and SEED for CBraMod), confirming that the improvements are consistent and practically meaningful rather than driven by performance gains on a subset of datasets.

\begin{table}[h]
\centering
\caption{Statistical comparison of EEG-PRIME against each baseline across benchmark datasets. W/L: number of datasets where EEG-PRIME outperforms / underperforms the baseline. Cohen's $d$ and one-sided
Wilcoxon signed-rank $p$-values treat each dataset as an independent observation.}
\label{tab:stat}
\resizebox{0.48\textwidth}{!}{%
\begin{tabular}{lcccl}
\toprule
Baseline    & W / L  & Cohen's $d$ & $p$-value & Losses on \\
\midrule
EEGNet      & 15 / 1 & 1.59 & $<$0.001 & OpenBMI-MI \\
Conformer   & 15 / 1 & 1.76 & $<$0.001 & OpenBMI-MI \\
TSception   & 15 / 1 & 0.95 & $<$0.001 & BCIC-Speech \\
STTrans.    & 16 / 0 & 1.86 & $<$0.001 & --- \\
BIOT        & 16 / 0 & 1.54 & $<$0.001 & --- \\
EEGPT       & 16 / 0 & 2.53 & $<$0.001 & --- \\
LaBraM      & 14 / 2 & 0.86 & 0.001    & SEED, BCIC-Speech \\
CBraMod     & 14 / 2 & 0.91 & $<$0.001 & SHU-MI, SEED \\
MIRepNet$^\dagger$ & 8 / 0  & 1.47 & 0.004 & --- \\
\bottomrule
\end{tabular}%
}
\\[4pt]
\footnotesize $^\dagger$MIRepNet is evaluated on 8 MI datasets only.
\end{table}
\subsection{Zero-Shot Inference}
\label{sec:zero_shot_inference}

\begin{figure*}[t]
\centering
\includegraphics[width=2\columnwidth]{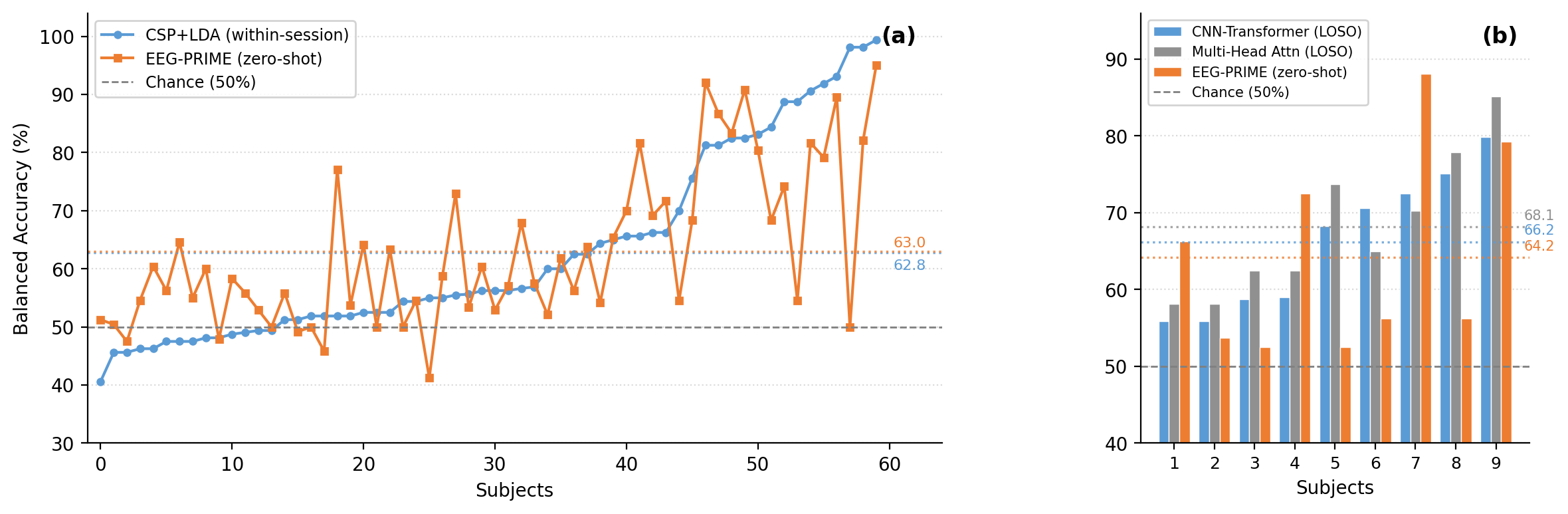}
\caption{Zero-shot inference results on two additional unseen datasets. After training, the model is evaluated directly on unseen target datasets without any additional fine-tuning, adaptation, or linear probing.}
\label{fig:zero_shot}
\end{figure*}

We evaluate EEG-PRIME in a fully zero-shot setting on two held-out datasets that were excluded from instruction tuning: Dreyer2023A (60 subjects) and Weibo2014 (10 subjects). For comparison, we include the within-session CSP+LDA results reported in the Dreyer2023A dataset release \citep{dreyer2023large}, where each subject's classifier is trained on two calibration runs (around 80 trials) from the same session, and LOSO-supervised baselines (CNN-Transformer and Multi-Head Attention \citep{otarbay2025svm}) on Weibo2014.
As shown in Fig.~\ref{fig:zero_shot}(a), EEG-PRIME achieves a mean balanced accuracy of 63.0\% across 60 subjects, closely matching the within-session CSP+LDA baseline (62.8\%~\citep{dreyer2023large}), despite using no subject-specific training data whatsoever. Notably, the per-subject accuracy profiles of the two methods exhibit substantial correlation ($r = 0.69$), indicating that EEG-PRIME captures genuine inter-subject variability in decoding discriminability rather than producing uniformly random predictions. Subjects with strong, well-defined neural representations are well-decoded by both methods, while subjects with weaker signals remain challenging for both.
On Weibo2014 (Fig.~\ref{fig:zero_shot}(b)), EEG-PRIME achieves a mean balanced accuracy of 64.2\% across nine subjects (one subject excluded as a known BCI-illiterate participant), compared to 66.2\% for CNN-Transformer and 68.1\% for Multi-Head Attention under the LOSO protocol. Considering that the supervised baselines leverage hundreds of labeled trials from the same dataset while EEG-PRIME relies solely on a natural language task description, this result demonstrates a remarkably small performance gap and underscores the practical viability of zero-shot EEG decoding.

\subsection{Ablation Studies}
\label{sec:ablation}

\begin{figure}[h]
    \centering
    \includegraphics[width=1.0\columnwidth]{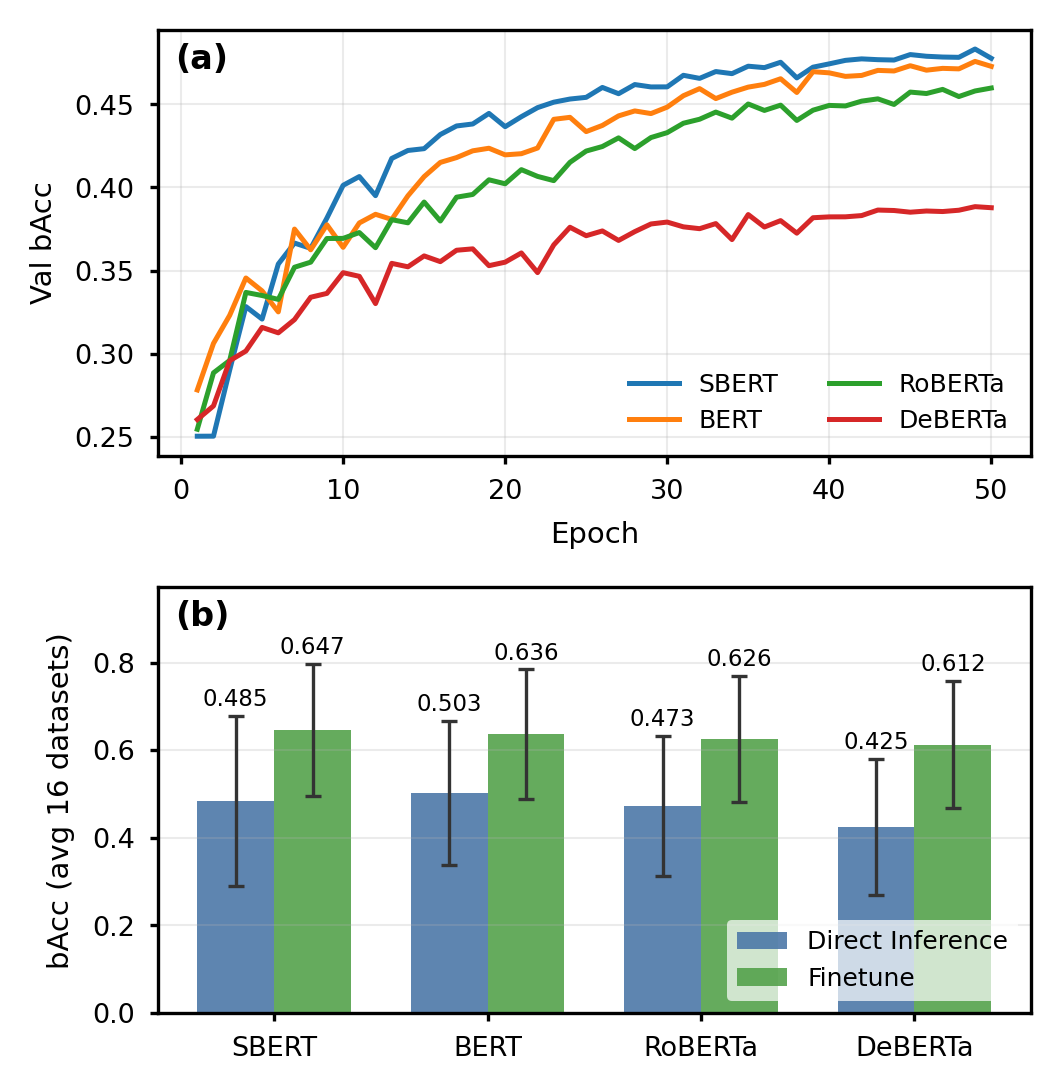}
    \caption{Comparison of text encoders during instruction tuning.
    \textbf{(a)} Instruction tuning validation balanced accuracy curves over 50 epochs for BERT, SBERT, RoBERTa, and DeBERTa.
    \textbf{(b)} Direct inference and fine-tuning balanced accuracy (B.Acc) averaged over all 16 downstream datasets.
    SBERT achieves the best convergence and highest fine-tuning performance, while BERT leads in direct inference.}
    \label{fig:language_models}
\end{figure}

\paragraph{Choice of Text Encoder}
The text encoder defines the semantic space into which EEG representations are aligned, making its choice a critical design decision.
We compare four frozen encoders from the BERT family: BERT-base-uncased \citep{devlin2019bert}, SBERT (all-mpnet-base-v2) \citep{reimers2019sentence}, RoBERTa \citep{liu2019roberta}, and DeBERTa \citep{he2020deberta}.
As shown in Fig.~\ref{fig:language_models}, SBERT converges fastest and reaches the highest instruction tuning validation accuracy by epoch 50, while DeBERTa converges most slowly and plateaus at the lowest accuracy.
Subfigure~(b) reports direct inference and in-domain fine-tuning results across all 16 datasets: BERT achieves the highest direct inference accuracy (B.Acc: 0.503), while SBERT achieves the highest fine-tuning performance (B.Acc: 0.647).
The full ranking by fine-tuning accuracy is SBERT (0.647) $>$ BERT (0.636) $>$ RoBERTa (0.626) $>$ DeBERTa (0.612), with corresponding direct inference values of 0.485, 0.503, 0.473, and 0.425, respectively.
These results confirm that SBERT (all-mpnet-base-v2), which is explicitly optimized for short sentence-level semantic similarity, provides the most effective semantic space for aligning EEG representations with task instructions and class-name labels, as evidenced by its superior convergence speed and highest fine-tuning performance.

\paragraph{Effect of Task Instructions and Dataset Embedding}
Table~\ref{tab:foundation16_prompt_ablation} summarizes an ablation on task instruction and the learned dataset embedding, using the same instruction-tuned checkpoint for both in-domain direct inference and downstream finetuning.
Specifically, we compare the full model, a variant that replaces all dataset-specific instructions with a shared default instruction (\textit{``default''}), and a variant that disables the learned dataset embedding while keeping the task-specific instruction.
During the ablation, the natural-language instructions and text prototypes are otherwise kept unchanged, so zeroing the dataset embedding directly tests whether the gain mainly comes from textual dataset descriptions alone.
In in-domain direct inference, replacing dataset-specific instructions with the default instruction yields a modest improvement (B.Acc: 0.4848$\to$0.5050), whereas removing the learned dataset embedding causes a substantial degradation (B.Acc: 0.4848$\to$0.4242; Kappa: 0.2179$\to$0.1111), indicating that direct transfer depends more strongly on the learned dataset embedding than on the precise wording of the task instruction.
In finetuning, the full model performs best, and removing either task-specific instructions or the dataset embedding leads to a comparable drop in performance (B.Acc: 0.6465$\to$0.6142 and 0.6465$\to$0.6165, respectively).
Overall, these results show that the dataset embedding is especially important in in-domain direct inference, while both forms of conditioning remain useful under finetuning.

\begin{table}[t]
\centering
\caption{Ablation of task instruction and learned dataset embedding on the downstream datasets. Results are averaged over all datasets. DS Emb.\ = dataset embedding; Finetune = task-specific fine-tuning.}
\label{tab:foundation16_prompt_ablation}
\renewcommand{\arraystretch}{0.9}
\begin{tabular}{lcc|cc}
\toprule
Task Instr. & DS Emb. & FT B.Acc / Kappa & In-domain B.Acc / Kappa \\
\midrule
\checkmark & \checkmark & \textbf{0.6465} / \textbf{0.4664} & 0.4848 / 0.2179  \\
          & \checkmark & 0.6142 / 0.4173 & \textbf{0.5050} / \textbf{0.2533} \\
\checkmark &           & 0.6165 / 0.4225 & 0.4242 / 0.1111  \\
\bottomrule
\end{tabular}
\end{table}

\subsection{Visualization of the Shared EEG–Language Embedding Space}
\begin{figure*}[t]
    \centering
    \includegraphics[width=1\linewidth]{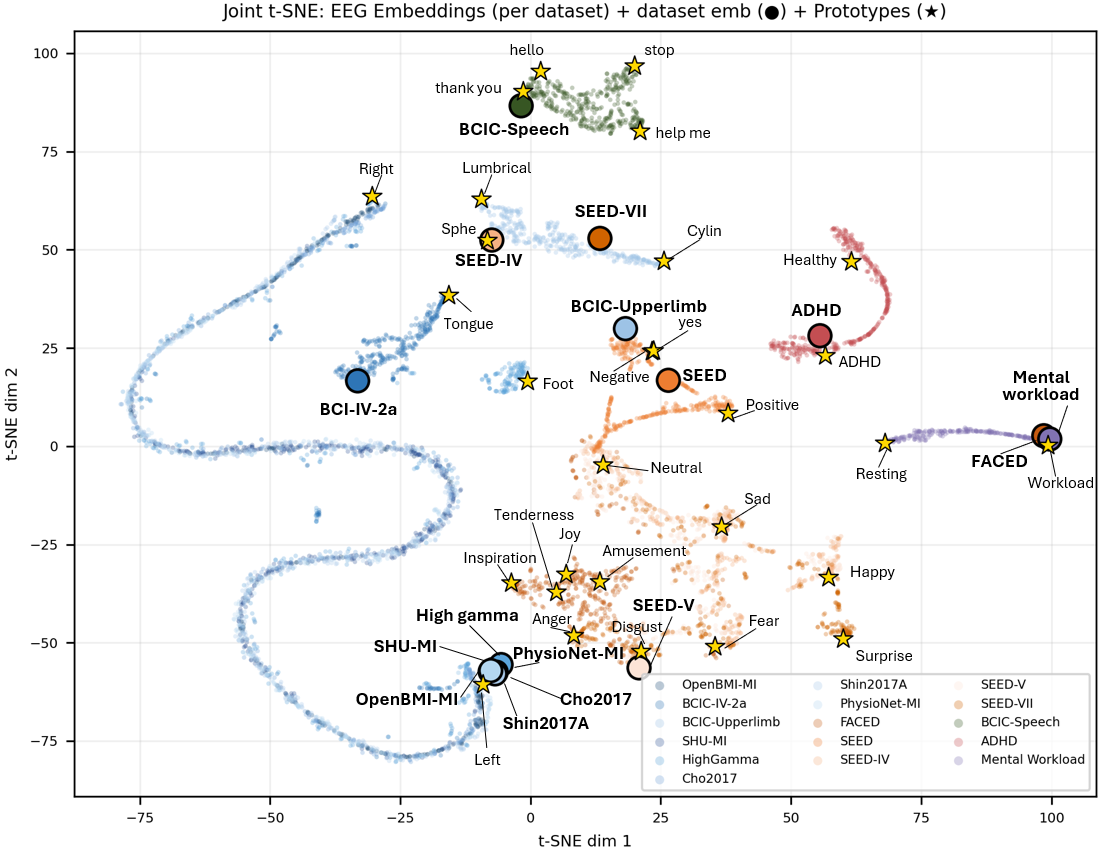}
    \caption{Joint t-SNE visualization of EEG embeddings and text prototypes 
    in the shared semantic space. Each point represents an EEG trial embedding 
    (colored by dataset); stars (\textbf{$\star$}) denote text prototype embeddings 
    of class labels; filled circles (\textbf{$\bullet$}) denote dataset conditioning 
    embeddings ($\mathbf{e}_d$). Embeddings from the same task domain cluster
    together (e.g., MI datasets form a contiguous manifold on the left), while 
    text prototypes of the corresponding classes are co-located within each 
    cluster, demonstrating cross-modal alignment between EEG representations 
    and natural language semantics without task-specific classification heads.}
    \label{fig:joint_tsne}
\end{figure*}

Fig.~\ref{fig:joint_tsne} visualizes the joint embedding space learned by EEG-PRIME via t-distributed stochastic neighbor embedding (t-SNE), projecting EEG trial embeddings from all 16 downstream datasets together with their corresponding text prototype embeddings. Several structural properties are evident. First, datasets sharing the same cognitive domain form coherent clusters: all eight MI datasets occupy a contiguous manifold in the left region of the space, while emotion recognition datasets (FACED, SEED variants) and task-specific datasets (ADHD, Mental Workload, Covert Speech) are distinctly separated. Second, text prototypes (stars) corresponding to class labels (such as \textit{Left}, \textit{Right}, \textit{Happy}, or \textit{ADHD}) are co-located with the EEG embeddings of their respective classes, confirming that the instruction-conditioned Q-Former successfully aligns neural signals with natural language semantics. Third, dataset conditioning embeddings ($\mathbf{e}_d$, filled circles)
are positioned near the centroid of their corresponding EEG cluster, suggesting that dataset-level conditioning provides a meaningful geometric anchor in the shared space. Together, these observations support the conclusion that EEG-PRIME learns a structured, semantically interpretable embedding space that generalizes across heterogeneous EEG tasks.

\subsection{Neurophysiological Interpretability: ERD--Attention Alignment}
\label{sec:erd_attn}

To assess whether the Q-Former's cross-attention reflects neurophysiologically meaningful signal, we examine the relationship between token-level attention weights and event-related desynchronization (ERD)—the well-established motor-imagery biomarker characterized by a suppression of mu/beta rhythms following movement onset.
Concretely, for each EEG token we compute its Q-Former attention probability (averaged over query slots and heads) and its mean ERD magnitude within the corresponding time window, then compare these two quantities across tokens and trials.

\begin{figure*}[t]
    \centering
    \includegraphics[width=2.0\columnwidth]{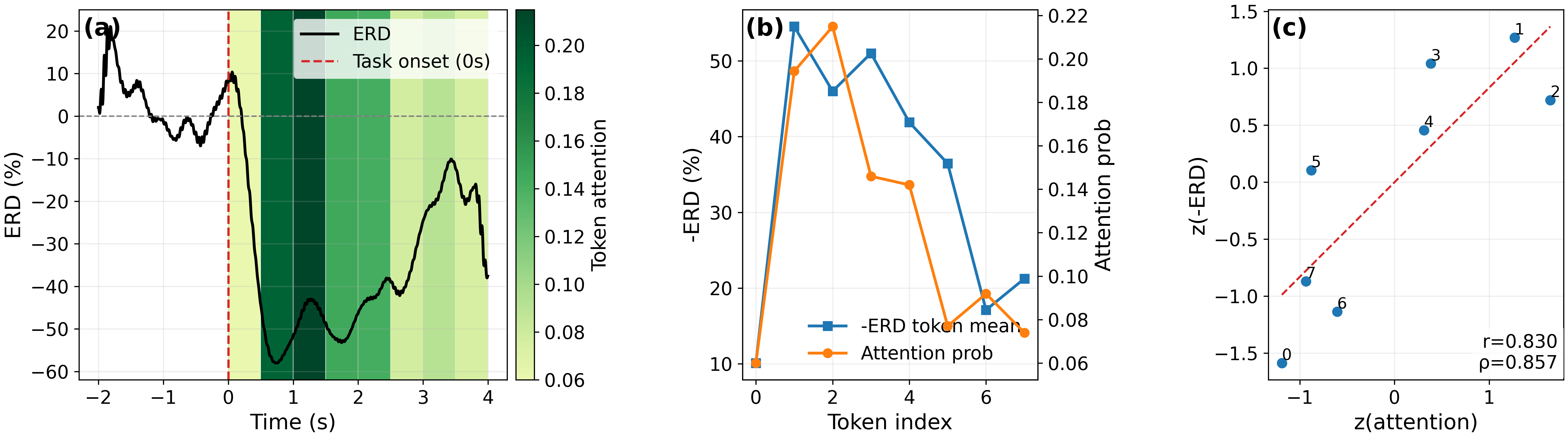}
    \caption{ERD--attention alignment of the Q-Former.
    \textbf{(a)} Trial-averaged ERD curve (black) with background shading proportional to per-token attention weight; darker green indicates higher attention.
    \textbf{(b)} Token-index profiles of mean $-$ERD (blue, left axis) and attention probability (orange, right axis).
    \textbf{(c)} Scatter plot of z-scored ERD vs.\ z-scored attention across token--trial pairs ($r{=}0.830$, $\rho{=}0.857$).}
    \label{fig:erd_attn}
\end{figure*}

Fig.~\ref{fig:erd_attn} summarizes the results.
Subfigure~(a) overlays the trial-averaged ERD curve with a color-coded background whose intensity encodes per-token attention weight: the Q-Former automatically assigns stronger attention to the post-onset interval (roughly 0.5--2\,s), precisely when ERD is most pronounced.
Subfigure~(b) plots token-index profiles for both -ERD (left axis) and attention probability (right axis); the two curves track each other closely across all eight token positions, indicating that the model's latent reader prioritizes tokens that carry stronger desynchronization.
Subfigure~(c) quantifies this co-variation with a scatter plot of z-scored ERD against z-scored attention across all token--trial pairs, yielding a Pearson correlation of $r = 0.830$ and a Spearman rank correlation of $\rho = 0.857$.

These results demonstrate that EEG-PRIME's Q-Former does not merely compress EEG signals mechanically but instead learns to selectively focus on time windows that are neurophysiologically relevant for MI decoding, providing evidence that the learned attention mechanism aligns with known EEG biomarkers.

\begin{figure*}[h]
    \centering
    \includegraphics[width=2.0\columnwidth]{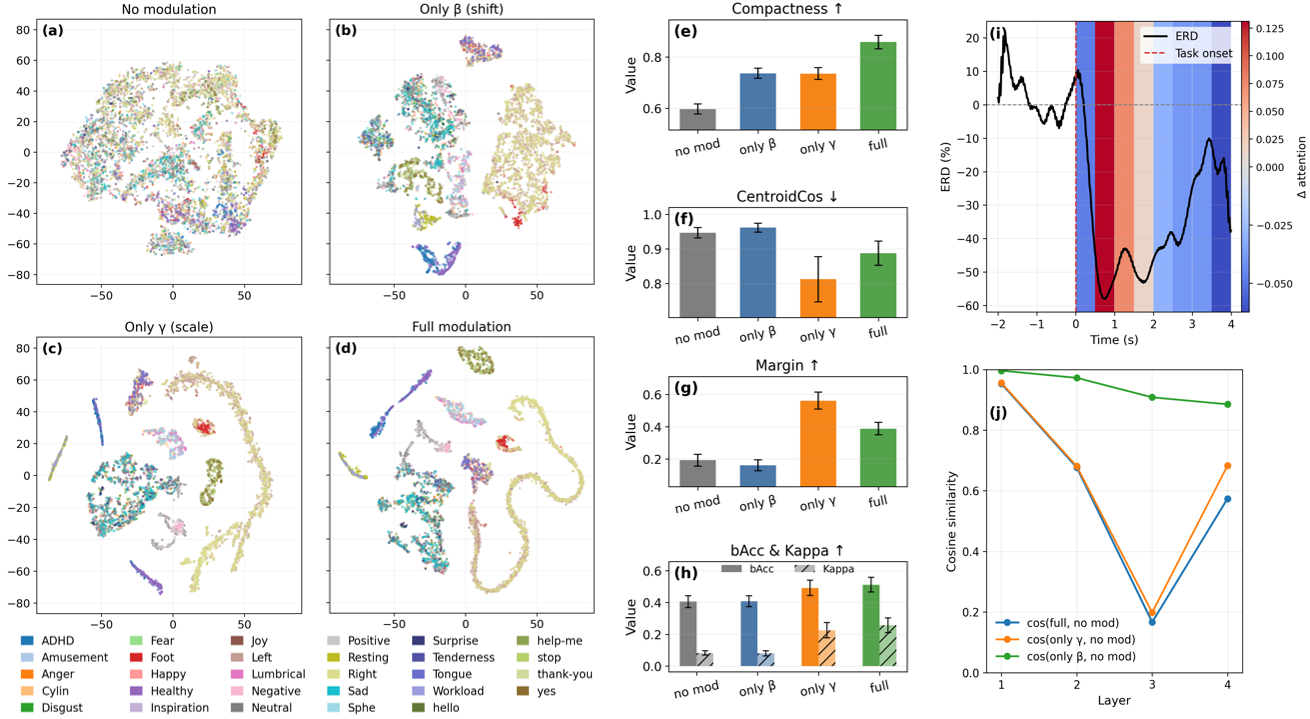}

\caption{Analysis of the LQM mechanism 
in the Q-Former across all 16 EEG decoding tasks.
\textbf{(a--d)} Joint t-SNE visualization of EEG trial embeddings under four LQM configurations (\textit{no modulation}, \textit{only-$\beta$} (shift only), \textit{only-$\gamma$} (scale only), and \textit{full modulation}), with points colored by semantic class label across all datasets. \textbf{(e--h)} Quantitative embedding-space metrics averaged across all 16 datasets (mean $\pm$ standard error of the mean, SEM): \textbf{(e)} intra-class compactness (mean cosine similarity to class centroid), \textbf{(f)} inter-class centroid cosine similarity (lower indicates greater separation), \textbf{(g)} decision margin (top-1 minus top-2 prototype similarity), and \textbf{(h)} balanced accuracy and Cohen's Kappa under in-domain direct inference. \textbf{(i)} Attention delta (modulated $-$ unmodulated) overlaid on the ERD curve for MI-BCIC-IV2a; red/blue background indicates increased/decreased cross-attention weight at each EEG token position.
\textbf{(j)} Per-layer cosine similarity of Q-Former hidden states between each modulated variant and the unmodulated baseline, showing the progressive divergence 
introduced by $\gamma$ and $\beta$.}
    \label{fig:counterfactual_qmod}
\end{figure*}

\section{Discussion}
\label{sec:discussion}

The quantitative results in Section~\ref{sec:main_results} demonstrate that EEG-PRIME consistently outperforms strong baselines across diverse BCI paradigms. In this section, we provide a deeper analysis of the mechanisms underlying these gains. We examine how LQM shapes the learned embedding geometry, analyze the relative contributions of its scale and shift components, and interpret the model's spatial attention patterns in terms of established electrophysiological signatures. We further visualize gradient-based saliency maps to assess the neurophysiological plausibility of the learned representations, and close with a discussion of current limitations and future directions.

We first investigate the causal role of LQM by performing a comparison between the full instruction-conditioned Q-Former and a modulation-ablated variant in which the LQM scale~($\gamma$) and shift~($\beta$) projections are disabled.

\subsection{Instruction Conditioning Reshapes the Embedding Geometry}

The t-SNE visualizations in Fig.~\ref{fig:counterfactual_qmod} (a)--(d) illustrate how progressively applying LQM components transforms the structure of the embedding space. 
Without any conditioning (subfigure a), EEG trial embeddings from semantically distinct classes form an undifferentiated manifold, with no discernible cluster boundaries across the 16 decoding tasks. Introducing the shift component $\beta$ alone (subfigure b) yields only marginal reorganization, leaving the overall topology largely unchanged. In contrast, activating the scale component $\gamma$ alone (subfigure c) induces a qualitative transition: class-specific embeddings begin to coalesce into compact, well-separated clusters, with linear and curved substructures emerging that reflect the underlying semantic categories. Full modulation ($\gamma + \beta$, subfigure d) further consolidates these structures, producing the most discriminative spatial arrangement. These observations suggest that $\gamma$, which multiplicatively rescales the query representations after layer normalization, acts as the primary geometric organizer in LQM, while $\beta$ plays a complementary but secondary role in refining cluster boundaries.
\subsection{Scale Modulation Dominates Quantitative Embedding Quality}

The quantitative metrics in Fig.~\ref{fig:counterfactual_qmod} (e)--(g) corroborate the qualitative 
trends observed in the t-SNE visualizations. Intra-class compactness (subfigure e), measured as the mean cosine similarity between each trial embedding and its class centroid, increases monotonically from \textit{no modulation} to \textit{full}, with \textit{only-$\gamma$} accounting for the majority of the improvement. Correspondingly, the mean pairwise cosine similarity between class centroids (subfigure f) decreases substantially under $\gamma$ modulation, indicating greater inter-class separation in the hyperspherical embedding space. The decision margin (subfigure g), defined as the difference between the top-1 and top-2 prototype cosine similarities, likewise increases markedly with $\gamma$, reflecting higher classification confidence. Notably, in some datasets the margin under \textit{only-$\gamma$} slightly exceeds that of \textit{full modulation}, a phenomenon attributable to the additive shift $\beta$: after L2 normalization, a uniform translational bias can compress angular distances between class centroids, partially offsetting the separative effect of $\gamma$. Across all three metrics, \textit{only-$\beta$} contributes negligibly beyond the unmodulated baseline, confirming that the scale modulation pathway is the functionally dominant component of LQM.

One possible explanation for this dominance is geometric in nature. After layer normalization, $\gamma$ performs a multiplicative rescaling of each activation dimension, which can rotate and stretch query vectors on the unit hypersphere and thereby alter which regions of the EEG token manifold each query attends to. In contrast, $\beta$ applies a uniform additive translation that shifts all queries by the same offset, preserving their relative angular relationships. Since cosine-similarity-based classification depends on the angular structure of the representation space rather than its absolute position, multiplicative rescaling may more effectively alter inter-class separability, while the additive component can only provide marginal adjustments at cluster boundaries. This asymmetry is broadly consistent with findings in instruction-conditioned generative models, where multiplicative modulation consistently shows stronger conditioning effect than additive shift, suggesting it may reflect a general tendency of adaptive layer normalization in transformer architectures.
\subsection{LQM Improves Downstream Classification Performance}

Figure~\ref{fig:counterfactual_qmod} (h) reports B.Acc and Cohen's Kappa averaged over 16 decoding tasks under in-domaindirect inference, with error bars denoting the SEM across datasets. Both metrics follow the same monotonic ordering as the geometric measures: \textit{only-$\gamma$} surpasses \textit{only-$\beta$} by a substantial margin, and \textit{full modulation} achieves the highest performance overall. While the additional gain from $\beta$ over \textit{only-$\gamma$} is modest in absolute terms, it is consistent across the majority of datasets (12 out of 16), suggesting that the additive shift systematically nudges borderline samples toward their correct class prototypes on the unit hypersphere, even without altering the global embedding geometry. The relatively large SEM reflects the inherent heterogeneity of the evaluation benchmark, spanning tasks that differ substantially in the number of classes, recording conditions, and subject populations. Nevertheless, the consistent ordering across all four configurations (\textit{no modulation} $<$ \textit{only-$\beta$} $<$ \textit{only-$\gamma$} $<$ \textit{full}) demonstrates that LQM provides a reliable and generalizable mechanism for aligning EEG representations with task-specific semantic structure.

\subsection{Attention Reallocation Aligns with Electrophysiological Signatures}

Figure~\ref{fig:counterfactual_qmod} (i) overlays the token-level attention delta (modulated minus unmodulated) on the ERD curve for an exemplar MI session (BCIC-IV2a). The attention reallocation induced by full modulation is not uniform across the trial: increased cross-attention weights (red) are preferentially concentrated in the post-stimulus window where ERD is most pronounced, whereas attention is suppressed (blue) during the pre-stimulus baseline. This systematic correspondence between the learned attention pattern and the canonical electrophysiological marker of MI suggests that LQM implicitly guides the Q-Former to attend preferentially to temporally informative EEG tokens. Crucially, this alignment emerges without any explicit supervision on attention allocation, indicating that the semantic content of the task instruction is sufficient to redirect the model's internal attention toward physiologically meaningful temporal segments of the EEG.
\subsection{Scale Modulation Drives Progressive Representational Divergence Across Layers}

Figure~\ref{fig:counterfactual_qmod} (j) traces the per-layer cosine similarity between the hidden query states of each modulated variant and the unmodulated baseline, providing a layer-wise view of how LQM reshapes the Q-Former's internal representations. The divergence introduced by $\gamma$ accumulates progressively across layers, reaching substantially larger values than $\beta$ at every depth. This trajectory reveals a fundamental asymmetry between the two modulation components: scale modulation persistently steers the representational pathway of the queries through the transformer stack, compounding its effect at each successive layer, whereas shift modulation induces only superficial perturbations that do not accumulate in the same manner. The fact that the representational gap between $\gamma$ and $\beta$ widens with network depth is consistent with the dominance of $\gamma$ in all downstream metrics reported in subfigures (e)--(h), and provides a mechanistic account of why scale modulation is the functionally critical component of LQM.

\subsection{Topography Visualization}
\label{sec:topography}

\begin{figure}[h]
    \centering
    \includegraphics[width=1.0\columnwidth]{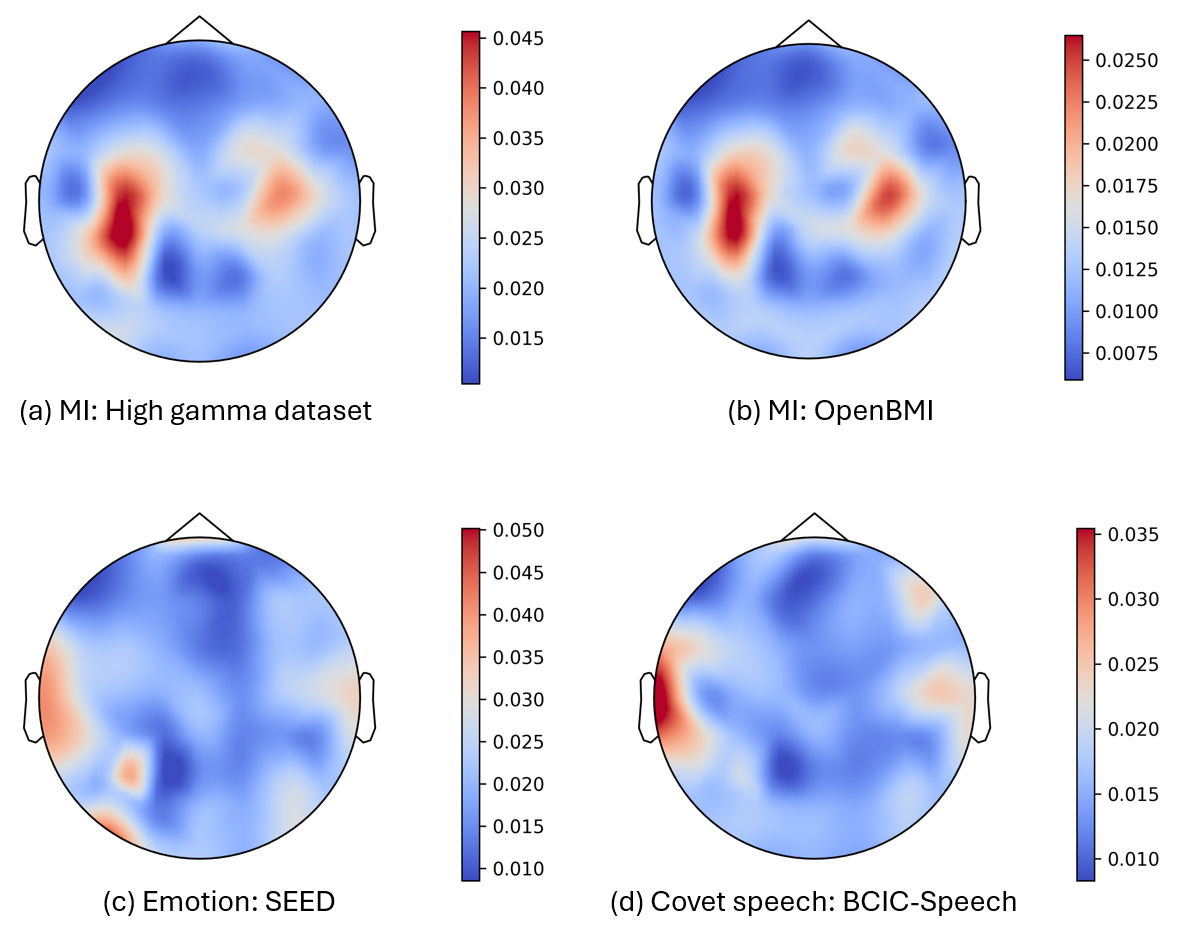}
    \caption{Saliency topomaps for representative tasks. Gradient-based saliency maps averaged over test samples, showing the mean absolute input gradient per electrode projected onto the scalp surface. }
    \label{fig:saliency}
\end{figure}

To interpret what spatial features the model relies on, we compute gradient-based saliency maps by backpropagating the score of the predicted class through the input EEG signal, retaining only correctly classified test samples to avoid noise from wrong predictions. Figure~\ref{fig:saliency} shows representative topomaps across four datasets. For MI (Fig~\ref{fig:saliency}(a)(b)), the model exhibits strong activation over the sensorimotor cortex (C3/C4), a well-established signature of ERD in the mu and beta bands \cite{zhang2022learning}. For emotion recognition (Fig~\ref{fig:saliency}(c)), salient regions are concentrated in the left temporal and parieto-occipital areas, consistent with prior work on the SEED benchmark reporting high discriminability of temporal-parietal channels \cite{ding2025emt}. For covert speech (Fig~\ref{fig:saliency}(d)), activation is left-lateralized over T7, FT7, and TP7, corresponding to Broca's and Wernicke's areas involved in speech production and comprehension \cite{jiang2026decoding}. These task-specific topographic patterns demonstrate that EEG-PRIME learns neurophysiologically meaningful spatial representations across heterogeneous EEG decoding tasks.

\subsection{Limitation and Future Work}
\label{sec:limitation}

Despite the encouraging zero-shot results on MI, zero-shot transfer is not uniformly successful across all paradigms. This limitation is expected because the difficulty of cross-dataset generalization varies substantially with label semantics, recording protocols, subject variability, and signal quality. In particular, tasks with larger distribution shifts or weaker shared neurophysiological structure, such as emotion recognition, remain more challenging under the strict zero-shot setting, where no target-domain calibration, adaptation, or linear probing is allowed. Therefore, our current results should be interpreted as evidence that EEG-PRIME makes zero-shot inference feasible for MI paradigms with clear neural correlates, rather than as a claim that zero-shot transfer is already solved universally.
This observation also suggests several directions for future work. First, more diverse pretraining corpora and richer task descriptions may improve cross-task transferability. Second, better alignment strategies between EEG features and textual class prototypes may further reduce dataset-specific mismatch.

\section{Conclusion}
We introduced EEG-PRIME, a multi-level conditioned EEG foundation model for cross-dataset multi-task EEG decoding. Built on the central hypothesis that language should guide how the model queries and interprets EEG rather than directly altering the neural signal, EEG-PRIME combines masked self-supervised pretraining with frequency-cutoff augmentation with a Q-Former conditioned on three types of prompts (task instructions, dataset instructions, and subject-invariance constraints) injected via LQM. Unified prediction across heterogeneous label spaces is achieved through text-anchored prototype classification, eliminating dataset-specific classifier heads. Across eighteen datasets spanning MI, emotion recognition, medical healthcare, covert speech, and mental workload (sixteen for task-specific fine-tuning and two datasets held out for zero-shot evaluation), EEG-PRIME demonstrates consistent improvements over strong EEG baselines and prior EEG foundation models under both zero-shot inference and dataset-specific fine-tuning.

\section*{Acknowledgments}
This research is supported by the Ministry of Education, Singapore, under its Academic Research Fund Tier 2 (Grant No. MOE-T2EP20124-0001).

\bibliographystyle{IEEEtran}
\bibliography{reference}

\end{document}